\documentclass[]{fairmeta}

\title{\vspace{-1pt}SoftVTBench: A Deformation-Aware Visuo-Tactile Dataset
and Benchmark for Deformable-Object Manipulation}

\author{
Bowen Jing$^{1,*}$,
Mingxin Wang$^{1,2,*}$,
Ruiyang Hao$^{3}$,
Chenchen Ge$^{1,4}$,
Hanwen Shen$^{5}$,
Junjie He$^{6}$,
Yang Cui$^{7}$,
Yiming Hou$^{1,4}$,
Weitao Zhou$^{2,8,\ddagger}$,
Jiawei Wang$^{8}$,
Minglei Li$^{8}$,
Dandan Zhang$^{9}$,
Ding Zhao$^{10}$,
Houde Liu$^{2}$,
Xiaofan Li$^{11}$,
Si Liu$^{12}$,
Ping Luo$^{13}$,
Haibao Yu$^{1,13,\ddagger}$
}

\affiliation[]{%
\noindent\footnotesize
$^{1}$Tuojing Intelligence, \quad $^{2}$Tsinghua University, \quad $^{3}$King's College London, \quad $^{4}$Southeast University, \\
$^{5}$Stevens Institute of Technology, \; $^{6}$The Hong Kong University of Science and Technology (Guangzhou), \\
$^{7}$University of Manchester, \quad $^{8}$Simple AI, \quad $^{9}$Imperial College London, \quad $^{10}$Carnegie Mellon University, \\
$^{11}$Zhejiang University, \quad $^{12}$Beihang University, \quad $^{13}$The University of Hong Kong
}

\contribution[*]{equal contribution}
\contribution[\ddagger]{corresponding author\vspace{-2ex}}

\usepackage{amsmath,amsfonts,bm}
\usepackage{xcolor}

\def\eqref#1{equation~\ref{#1}}

\def\1{\bm{1}}

\DeclareMathAlphabet{\mathsfit}{\encodingdefault}{\sfdefault}{m}{sl}
\SetMathAlphabet{\mathsfit}{bold}{\encodingdefault}{\sfdefault}{bx}{n}

\usepackage{graphicx}
\usepackage{xcolor}
\usepackage{colortbl}

\usepackage{pifont}
\usepackage{makecell}
\usepackage{booktabs}
\usepackage{tabularx}
\usepackage{array}
\usepackage{multirow}
\usepackage{diagbox}
\usepackage{hhline}
\usepackage{longtable}
\usepackage{siunitx}
\usepackage{adjustbox}
\usepackage{wrapfig}

\definecolor{okgreen}{RGB}{30,132,73}
\definecolor{nored}{RGB}{192,57,43}
\definecolor{neutralgray}{RGB}{127,140,141}
\definecolor{ourshl}{RGB}{235,240,245}
\definecolor{partialorange}{RGB}{213,94,0}
\definecolor{lightgray}{rgb}{0.95,0.95,0.95}
\definecolor{baselinecolor}{gray}{.9}

\newcommand{\cmark}{\textcolor{okgreen}{\ding{51}}}
\newcommand{\xmark}{\textcolor{nored}{\ding{55}}}
\newcommand{\pmark}{\textcolor{partialorange}{\ding{109}}}

\newcolumntype{C}{>{\centering\arraybackslash}X}

\usepackage{amsmath,amsfonts,amssymb}
\usepackage{bm}
\usepackage{nicefrac}

\usepackage{enumitem}
\setlist[itemize]{leftmargin=*}

\usepackage{caption}
\crefname{figure}{Figure}{Figures}
\crefname{table}{Table}{Tables}

\usepackage{xspace}
\usepackage{calc}
\usepackage{etoolbox}
\usepackage{fancyvrb}
\usepackage{tikz}

\usepackage{titletoc}

\titlecontents{section}
[1.5em]
{\addvspace{-0.5pt}}
{\bfseries\contentslabel{2.3em}}
{\hspace*{-2.3em}\bfseries}
{\bfseries\titlerule*[.5pc]{.}\contentspage}
\titlecontents{subsection}
[3.8em]
{\addvspace{-2.2pt}}
{\contentslabel{2.3em}}
{\hspace*{-2.3em}}
{\titlerule*[.5pc]{.}\contentspage}

\abstract{
Physical interaction quality is central to deformable-object
manipulation, yet most benchmarks evaluate task success alone. A policy
may complete the task while allowing slip or causing excessive
compression. A primary bottleneck is the absence of visuo-tactile
datasets that pair policy-visible contact observations with independent
physical ground truth over complete tasks. We introduce
\textbf{SoftVTBench}, a visuo-tactile dataset for
physical-interaction-aware deformable-object manipulation. It contains
$4{,}000$ expert demonstrations and more than $50$ assets, including volumetric
deformable objects and visually matched rigid twins. At $20$\,Hz, each
episode synchronizes multi-view RGB, dual-finger tactile RGB and marker
motion, proprioception, language, and binary and continuous gripper
actions, alongside evaluator-only finite-element (FEM) states. Building
upon this dataset, we establish a closed-loop benchmark that uses fixed
object-specific calibration to define the
\textbf{Deformation-aware Success Rate (DSR)}, which counts a rollout as
successful only when it completes the task and keeps peak normalized
deformation within tolerance. Across Diffusion Policy, $\pi_{0.5}$, and
FastWAM, all $12$ in-distribution configurations contain successful
rollouts that violate the deformation tolerance, accounting for
$0.7$--$24\%$ of each configuration's successes. Under distribution
shift, visuo-tactile variants achieve higher task success in all six
policy--suite comparisons and higher DSR in five, whereas their
in-distribution benefits are mixed. These results show that making touch
available does not by itself ensure effective multimodal fusion.
SoftVTBench therefore provides a common visuo-tactile resource for
studying not only whether a policy succeeds, but how it physically
interacts with deformable objects and when touch improves that
interaction.
\vspace{-10pt}
}

\metadata[Code]{\url{https://github.com/TuojingAI/SoftVTBench}}
\metadata[Website]{\url{https://softvtbench.github.io/}}

\begin{document}
\maketitle

\section{Introduction}
\label{sec:intro}

\begin{figure*}[t]
    \centering
    \includegraphics[width=1\linewidth]{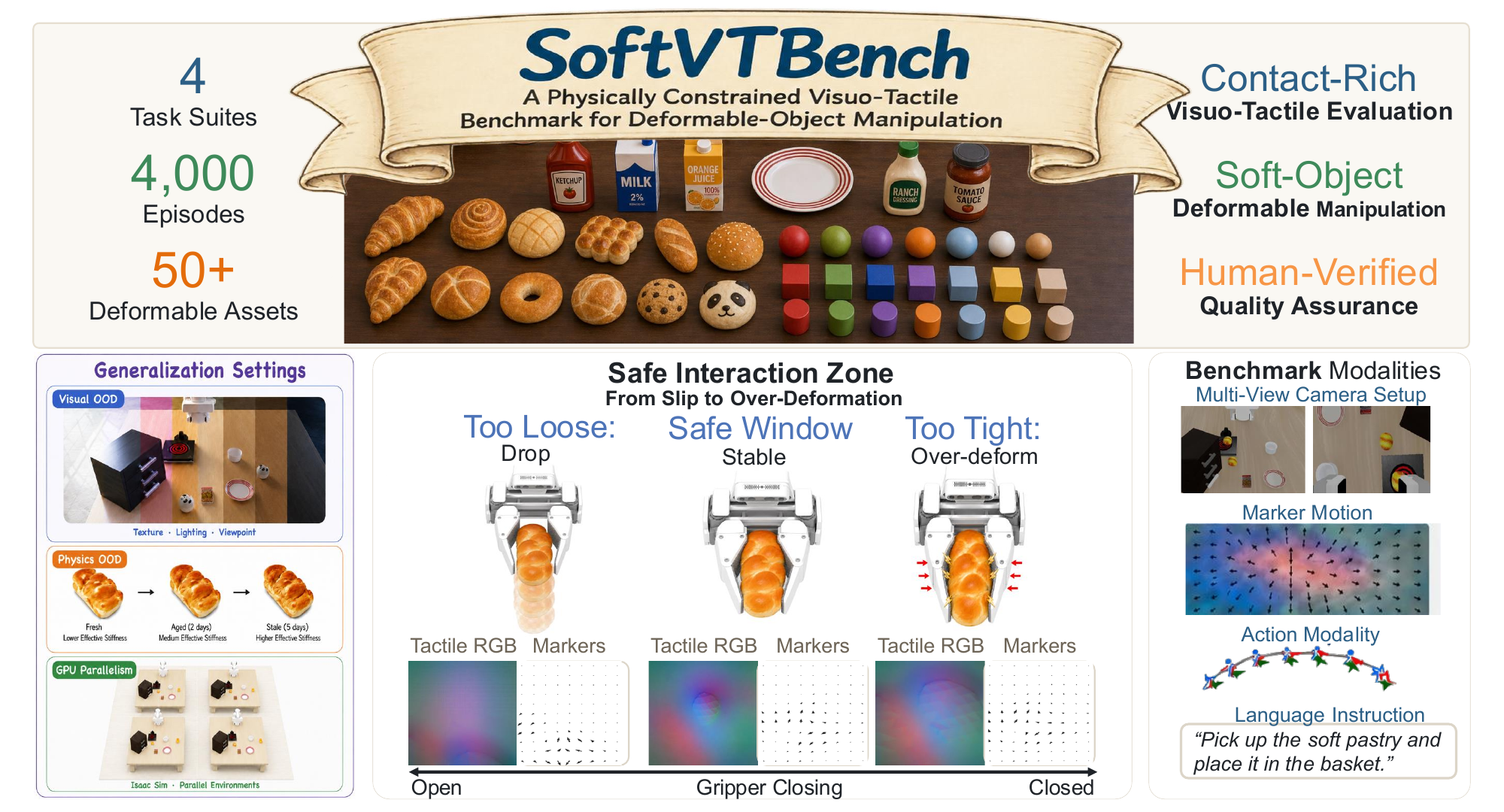}
    \caption{
    \textbf{Overview of SoftVTBench.}
    \textbf{Top:} $4{,}000$ demonstrations over four diagnostic suites
    and more than $50$ assets, including volumetric deformable objects
    and visually matched rigid twins.
    \textbf{Bottom left:} controlled visual and physical shifts for
    closed-loop generalization evaluation.
    \textbf{Bottom center:} the physical-interaction regimes captured by
    touch and evaluated from object deformation, from slip-prone loose
    grasps to excessive compression.
    \textbf{Bottom right:} synchronized visual, tactile,
    proprioceptive, language, and action streams.
    }
    \label{fig:overview}
\end{figure*}

Robotic manipulation benchmarks define what counts as progress. Existing
suites have standardized task diversity, language grounding, and
generalization~\cite{liu2023libero,nasiriany2024robocasa,gu2024simpler,mu2025robotwin},
but most evaluate whether a policy completes the instructed task. Task
success is necessary, yet it does not describe the physical interaction
that produced the outcome. Closing this gap requires more than a new
metric. The missing component is not a metric but a record: no existing dataset exposes contact sensing to the policy while independently logging the physical consequences of that contact.

Deformable-object manipulation makes this limitation concrete. A policy
may fail because a loose grasp allows the object to slip, but it may also
complete the task by compressing the object beyond an acceptable range.
Task success distinguishes neither failure mode from a gentle, stable
interaction. This distinction matters for food, soft packaging, medical
materials, and other deformation-sensitive objects~\cite{zhu2022challenges}.
Vision supplies the scene context needed to approach the object and
reach the target, but the gripper--object interface becomes partly
occluded precisely when contact begins. Touch complements vision at this
stage: tactile images and marker motion capture contact geometry, shear,
slip, and local compression directly at the interface
~\cite{yuan2017gelsight,suresh2024neuralfeels}. A visuo-tactile record
therefore describes both where the manipulation occurs and how contact
evolves while it is executed.

This creates a specific dataset requirement: one recording must carry two
disjoint streams. The policy-visible stream synchronizes vision, touch, robot
state, language, and actions, so that contact-aware behavior can be learned
over a complete trajectory rather than at an isolated grasp. The
evaluator-only stream records the object's own physical state and must remain
hidden from the policy, since tactile observations cannot serve as their own
evaluation target. Touch tells the policy how contact is evolving; the hidden
physical state tells the evaluator what that contact did to the object.

Existing benchmarks provide individual parts of this loop, but not the full
combination. General manipulation suites emphasize complete tasks and
generalization but mainly use rigid objects and task-success
evaluation~\cite{liu2023libero,nasiriany2024robocasa,gu2024simpler,mu2025robotwin}.
Deformable-object benchmarks model soft-body dynamics but often make
deformation the task objective, evaluate isolated grasps, or omit tactile
observations~\cite{lin2020softgym,huang2021plasticinelab,greenland2024sograb,zhang2025modesuite}.
Visuo-tactile benchmarks expose contact sensing to the policy but do not
provide evaluator-only deformation ground truth over complete
tasks~\cite{luu2025manifeel,zorin2026taco,Liu2025VTDexManip}. As
Table~\ref{tab:positioning} summarizes, no existing benchmark jointly provides
complete manipulation tasks, volumetric deformable objects, a policy-visible
tactile stream, and evaluator-only physical states that make deformation part
of the evaluation criterion.

To fill this gap, we introduce \textbf{SoftVTBench}, a visuo-tactile dataset for physical-interaction-aware deformable-object manipulation (Figure~\ref{fig:overview}). It contains
$4{,}000$ expert demonstrations over $40$ pick-and-place tasks and more
than $50$ assets, including volumetric deformable objects and visually
matched rigid twins. In every trajectory, multi-view RGB provides scene
and task context, while dual-finger tactile RGB and marker motion record
contact appearance and local shear. These streams are synchronized with
proprioception, language, and arm and gripper actions at $20$\,Hz. The
same interaction is paired with evaluator-only FEM, object, and contact
states, making the dataset useful both for visuo-tactile policy learning
and for auditable physical-interaction evaluation.

Building upon this visuo-tactile dataset, we establish a closed-loop benchmark
for physical-interaction evaluation. We obtain the hidden physical state from a
finite-element (FEM) representation that models each deformable object as a
volumetric mesh and tracks its nodes under contact; after removing rigid-body
transport, the residual nodal displacement measures how much the object changed
shape during the rollout. Before any policy is trained, scripted physical
probing sweeps the gripper closure over this measure to fix an object-specific
deformation tolerance, so the criterion is calibrated independently of the
policies it scores. During evaluation, the policy receives only the
policy-visible stream, whereas the evaluator reads the FEM states and records
peak normalized deformation over the rollout. The resulting \textbf{Task
Success Rate (TSR)} and \textbf{Deformation-aware Success Rate (DSR)} separate
completion from interaction quality: DSR credits an episode only when it both
completes the task and stays within the calibrated tolerance, so their
difference is exactly the share of successes that were physically unsafe.
Matched rigid twins, dual gripper-action encodings, and paired
in-distribution (ID) and out-of-distribution (OOD) protocols support
controlled analysis of deformability, contact sensing, control granularity,
and generalization.

We evaluate Diffusion Policy, $\pi_{0.5}$, and FastWAM under paired
vision-only and visuo-tactile settings~\cite{chi2023diffusion,physical2025pi,yuan2026fastwam}.
Across all $12$ in-distribution deformable-object configurations, DSR
identifies rollouts that complete the task while exceeding the calibrated
deformation tolerance. The results further show that physical-interaction
performance differs across policy families, that apparent tactile gains
can be confounded by gripper control granularity, and that touch is most
consistently associated with stronger results under distribution shift.
These findings
also show that making tactile observations available does not guarantee
that a policy will fuse them effectively. They serve as evidence for the
benchmark's necessity and diagnostic value, rather than as a claim that
touch is uniformly beneficial.

In summary, we make three contributions:
\begin{itemize}\setlength{\itemsep}{1pt}
    \item We introduce a visuo-tactile dataset comprising $4{,}000$
    expert demonstrations of deformable-object manipulation across $40$
    tasks in four diagnostic suites. Each trajectory pairs multi-view
    RGB with dual-finger tactile RGB and marker motion, synchronized with
    proprioception, language, actions, and evaluator-only physical states
    over matched rigid--deformable assets.
    \item We establish a closed-loop benchmark centered on DSR, which
    incorporates task success and FEM-grounded deformation compliance
    through policy-independent, object-specific calibration. Matched
    rigid twins and ID/OOD protocols support controlled physical
    comparisons.
    \item Across three policy families, we demonstrate that task success
    alone conceals excessive deformation, and use the benchmark's matched
    controls to analyze the effects of policy family, deformability,
    tactile sensing, gripper control granularity, and distribution shift.
\end{itemize}

\section{Related Work}
\label{sec:related-work}

\begin{table*}[t]
\centering
\footnotesize
\renewcommand{\arraystretch}{1.15}
\begin{tabular*}{\textwidth}{@{\extracolsep{\fill}}
l
>{\centering\arraybackslash}p{1.8cm}
>{\centering\arraybackslash}p{2.0cm}
>{\centering\arraybackslash}p{2.2cm}
>{\centering\arraybackslash}p{2.8cm}
@{}}
\toprule
\textbf{Benchmark}
& \textbf{Complete Task}
& \textbf{3D Deformable}
& \textbf{Policy-Visible Touch}
& \textbf{Evaluator-Only Deformation Scoring} \\
\midrule
LIBERO~\cite{liu2023libero}
& \cmark & \xmark & \xmark & \xmark \\
ManiSkill2~\cite{gu2023maniskill2}
& \cmark & \cmark & \xmark & \xmark \\
SoftGym~\cite{lin2020softgym}
& \cmark & \pmark & \xmark & \xmark \\
MoDeSuite~\cite{zhang2025modesuite}
& \cmark & \pmark & \xmark & \xmark \\
DefGraspSim~\cite{huang2022defgraspsim}
& \xmark & \cmark & \xmark & \cmark \\
SoGraB~\cite{greenland2024sograb}
& \xmark & \cmark & \xmark & \cmark \\
VTDexManip~\cite{Liu2025VTDexManip}
& \cmark & \xmark & \cmark & \xmark \\
ManiFeel~\cite{luu2025manifeel}
& \cmark & \xmark & \cmark & \xmark \\
Tabero~\cite{wu2026tabero}
& \cmark & \xmark & \cmark & \pmark \\
\midrule
\rowcolor{ourshl}
\textbf{SoftVTBench}
& \cmark & \cmark & \cmark & \cmark \\
\bottomrule
\end{tabular*}
\caption{
\textbf{Benchmark positioning.}
\emph{Complete Task}: the benchmark scores multi-stage manipulation
(approach, grasp, transport, place) rather than an isolated grasp.
\emph{3D Deformable}: the manipulated objects are volumetric soft
bodies, as opposed to cloth or rope (\pmark).
\emph{Policy-Visible Touch}: tactile observations are available to the
policy as input.
\emph{Evaluator-Only Deformation Scoring}: object deformation is
measured from physical state hidden from the policy, and that
measurement enters the success criterion, so an episode can complete
the task and still be scored as a failure; \pmark{} marks benchmarks
that constrain interaction on the input side (e.g.\ a force budget)
without scoring the resulting deformation.
Symbols denote full (\cmark), partial (\pmark), or no (\xmark)
support. Benchmarks that expose touch do not score deformation from
hidden state, and those that do evaluate isolated grasps; only
SoftVTBench provides all four.
}
\label{tab:positioning}
\end{table*}

\paragraph{Robotic Manipulation Benchmarks.}
General-purpose manipulation benchmarks standardize complete-task
evaluation across multi-task, language-conditioned, and generalization
settings. Representative suites include Meta-World~\cite{yu2020metaworld},
RLBench~\cite{james2020rlbench}, LIBERO~\cite{liu2023libero},
CALVIN~\cite{mees2021calvin}, RoboCasa~\cite{nasiriany2024robocasa},
RoboTwin~\cite{mu2025robotwin, chen2025robotwin2}, THE
COLOSSEUM~\cite{pumacay2024colosseum}, and
SIMPLER~\cite{gu2024simpler}. These benchmarks primarily measure task
success and therefore do not reveal the physical consequences of
contact. Safety-oriented benchmarks~\cite{zhang2025safevla,
fan2026safevlabench, huang2026safemanip} additionally evaluate
constraint violations and semantic hazards, but still focus on rigid
objects. Neither line directly evaluates whether a policy completes a
deformable-object task without excessive deformation.

\paragraph{Deformable-Object Manipulation.}
Deformable-object benchmarks provide the physical models needed to
represent shape change~\cite{zhu2022challenges}. SoftGym~\cite{lin2020softgym},
DEDO~\cite{antonova2021dedo}, and GarmentLab~\cite{lu2024garmentlab}
cover cloth, rope, and garment manipulation. PlasticineLab~\cite{huang2021plasticinelab}
and DaXBench~\cite{chen2022daxbench} provide differentiable soft-body
simulation, while ManiSkill2~\cite{gu2023maniskill2} and
MoDeSuite~\cite{zhang2025modesuite} place deformable objects within
broader manipulation suites. Recent foundation-model policies further
study generalization in these settings~\cite{su2026demavla}. In many of
these tasks, however, changing the object's shape is the objective.
SoftVTBench studies the complementary gentle-handling regime, in which
a policy must complete a separate task while keeping deformation
within a calibrated tolerance. Closest to this setting,
SoGraB~\cite{greenland2024sograb}, DefGraspSim~\cite{huang2022defgraspsim},
and DefGraspNets~\cite{huang2023defgraspnets} measure grasp-induced
deformation or stress from physical simulation. Their evaluation is
limited to isolated grasps and does not test whether touch helps a
closed-loop policy preserve an object throughout a complete task.

\paragraph{Visuo-Tactile Sensing and Policy Learning.}
Tactile sensing matters for gentle handling because it
captures contact geometry, shear, slip, and compression that are
difficult to infer from vision alone~\cite{yuan2017gelsight,
suresh2024neuralfeels}. Simulators including Taxim~\cite{si2022taxim},
FOTS~\cite{zhao2024fots}, TacEx~\cite{nguyen2024tacex},
TacSL~\cite{akinola2025tacsl}, and DiffTactile~\cite{si2024difftactile}
have supported tactile policy learning. Recent vision-language-action
models also incorporate touch, including VTLA~\cite{zhang2025vtla},
OmniVTLA~\cite{cheng2025omnivtla}, VLA-Touch~\cite{bi2025vlatouch},
Tactile-VLA~\cite{huang2025tactilevla}, and
AT-VLA~\cite{li2026atvla}. ManiFeel~\cite{luu2025manifeel},
TacO~\cite{zorin2026taco}, and VTDexManip~\cite{Liu2025VTDexManip}
further study visuo-tactile policy learning, with VTDexManip focusing
on dexterous manipulation of rigid objects. Together, these works show
how to expose tactile information to a policy, but they typically
evaluate its benefit through task success in rigid or incidentally
deformable settings.

Closest in spirit, Tabero~\cite{wu2026tabero} evaluates
language-conditioned gentle manipulation with closed-loop force
feedback. It scores interaction quality alongside task success, but
uses contact force on rigid objects rather than measured deformation
of a soft body. We therefore mark it as partial deformation-aware
evaluation in Table~\ref{tab:positioning}. Among the benchmarks
compared in the table, none jointly provides complete manipulation
tasks, volumetric deformable objects, policy-visible touch, and
evaluator-only physical ground truth. SoftVTBench combines these
elements, and its Deformation-aware Success Rate (DSR) incorporates
task success and compliance with a calibrated deformation tolerance.
This design makes it possible to test when visuo-tactile policies
improve physical interaction, rather than task success alone.

\section{The SoftVTBench Dataset}
\label{sec:dataset}

SoftVTBench is a visuo-tactile dataset for learning and evaluating
contact-rich deformable-object manipulation. Its policy-facing core
pairs multi-view RGB, which provides scene and task context, with
dual-finger tactile RGB and marker motion, which directly observe the
contact interface. It contains $4{,}000$ expert demonstrations across
$40$ pick-and-place tasks in four diagnostic suites, together with more
than $50$ assets. Every demonstration synchronizes these visuo-tactile
observations with proprioception, language, and actions, while recording
evaluator-only FEM, object, and contact states on the same timeline. The
dataset therefore supports both visuo-tactile policy learning and
independent deformation-aware evaluation. SoftVTBench is implemented in
Isaac Sim~\cite{NVIDIA_Isaac_Sim} and Isaac
Lab~\cite{mittal2025isaaclab}.

\begin{figure*}[t]
    \centering
    \includegraphics[width=1\linewidth]{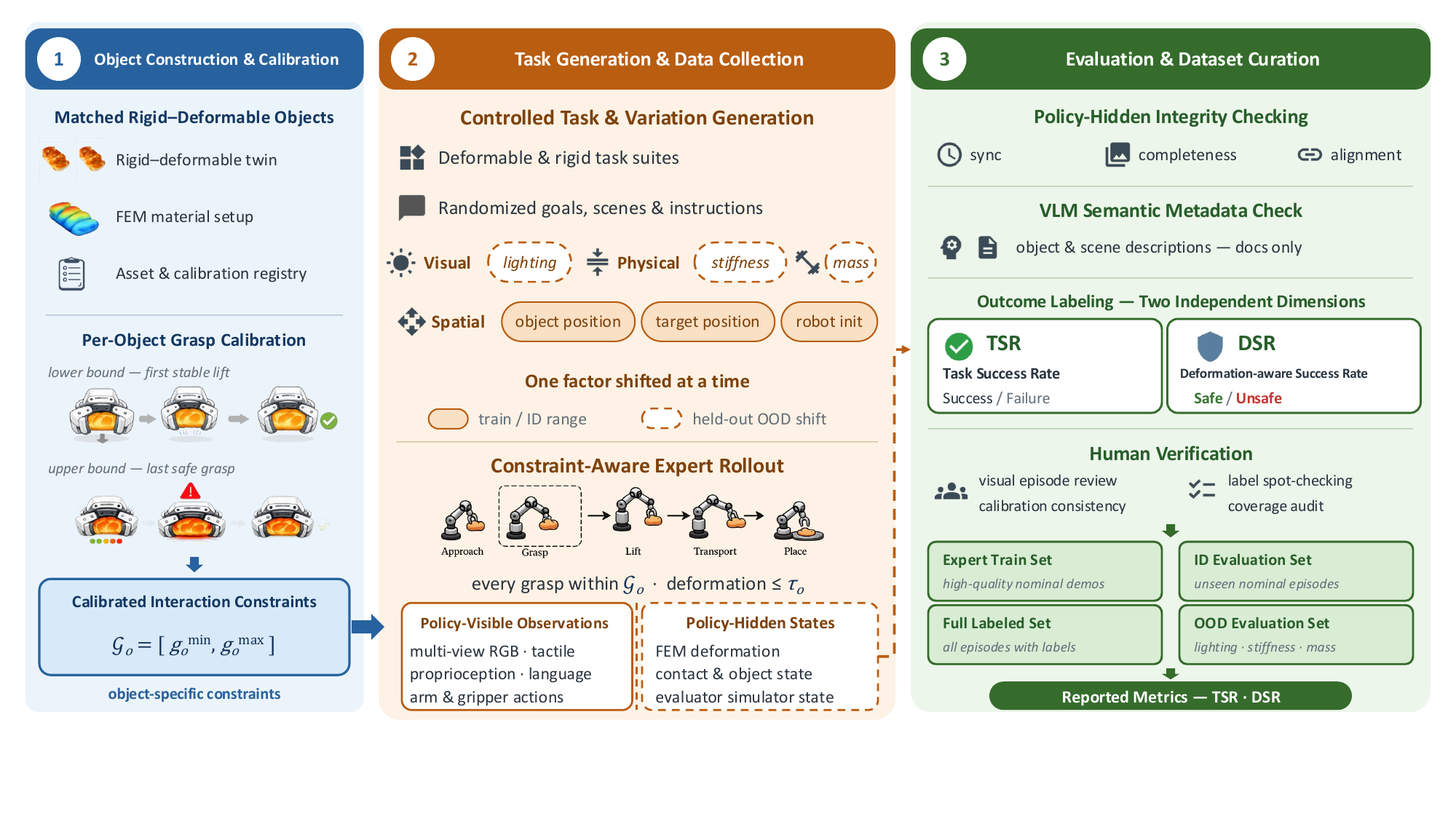}
    \caption{
    \textbf{SoftVTBench construction and evaluation pipeline.}
    Stage 1 constructs matched rigid--deformable objects and calibrates
    object-specific interaction constraints. Stage 2 generates
    controlled tasks and records policy-visible observations separately
    from evaluator-only physical states. Stage 3 applies automatic
    quality control and human verification, producing the released
    training, ID, and OOD splits. Task success and deformation compliance
    are labeled separately; DSR combines them during benchmark evaluation.
    }
    \label{fig:pipeline}
\end{figure*}

Figure~\ref{fig:pipeline} summarizes how the released resource is built.
Matched assets and controlled variations provide diagnostic comparisons,
while synchronized policy-visible and evaluator-only streams record two
views of the same physical interaction. Expert demonstrations are then
screened for integrity and outcome consistency before being assigned to
the training and evaluation splits. The remainder of this section first
describes the released data and then documents task generation,
collection, curation, and per-object calibration.

\subsection{Dataset Composition}
\label{sec:composition}

\paragraph{Tasks, Objects, and Scenes.}
The demonstrations are evenly divided among \textsc{Object-Soft},
\textsc{Spatial-Soft}, \textsc{Object-Rigid}, and
\textsc{Spatial-Rigid}, with $10$ tasks and $1{,}000$ demonstrations per
suite (Table~\ref{tab:dataset_stats}). The dataset contains $10$
volumetric deformable assets, including naturalistic bakery-style objects
and procedurally generated primitives, covering variation in geometry,
compliance, contact area, and appearance. Each deformable object has a
rigid twin
with matched mesh, texture, and mass but negligible grasp-induced
deformation. Twin tasks reuse the corresponding layout and language
instruction. Together with tables, floors, cabinets, containers, and
rigid distractors, the tabletop scenes contain more than $50$ assets in
total. Placement regions are adapted from LIBERO~\cite{liu2023libero},
and the appendix provides deformable-asset details and the task list.

\begin{table*}[t]
\centering
\renewcommand{\arraystretch}{1.15}
\setlength{\tabcolsep}{6pt}
\small
\begin{tabular}{@{}llccccc@{}}
\toprule
\textbf{Suite} & \textbf{Object Type} & \textbf{Variation Axis} &
\textbf{\#Tasks} & \textbf{\#Demos} &
\makecell[c]{\textbf{ID Eval}\\\textbf{Episodes}} &
\makecell[c]{\textbf{OOD}\\\textbf{Conditions}} \\
\midrule
\textsc{Object-Soft}   & Deformable & Object identity & 10 & 1{,}000 & 500 & 9 \\
\textsc{Spatial-Soft}  & Deformable & Spatial layout  & 10 & 1{,}000 & 500 & 9 \\
\textsc{Object-Rigid}  & Rigid twin & Object identity & 10 & 1{,}000 & 500 & -- \\
\textsc{Spatial-Rigid} & Rigid twin & Spatial layout  & 10 & 1{,}000 & 500 & -- \\
\midrule
\textbf{Total} & -- & -- & \textbf{40} & \textbf{4{,}000} & \textbf{2{,}000} & -- \\
\bottomrule
\end{tabular}
\normalsize
\caption{
\textbf{Dataset composition and evaluation splits.}
The four suites form a matched $2\times2$ design over object type and
variation axis. OOD evaluation uses nine single-factor conditions on
the two deformable suites.
}
\label{tab:dataset_stats}
\end{table*}

\paragraph{Dataset Splits.}
The accepted demonstrations form the supervision set. In-distribution
evaluation uses $500$ held-out initial states per suite, $50$ per task,
whose variation factors remain within the training support.
Out-of-distribution evaluation covers nine single-factor conditions
applied to the two deformable suites: dome-light intensity moves from
$135$ to $180$, $270$, and $67.5$; object mass is scaled by $1.25$,
$1.75$, and $2.5$; and Young's modulus is scaled by $0.8$, $0.5$, and
$2.0$. Each condition changes one factor outside its training support
while holding the remaining task conditions fixed, and reuses the task,
initial state, and seed of its in-distribution reference so that the
perturbed parameter is isolated.

\subsection{Recorded Streams}
\label{sec:streams}

Every episode records policy-visible observations and evaluator-only
physical states on the same $20$\,Hz timeline. Their separation allows
a policy to use touch without accessing the ground truth used to score
its physical interaction.

\begin{figure*}[t]
    \centering
    \includegraphics[width=\linewidth]{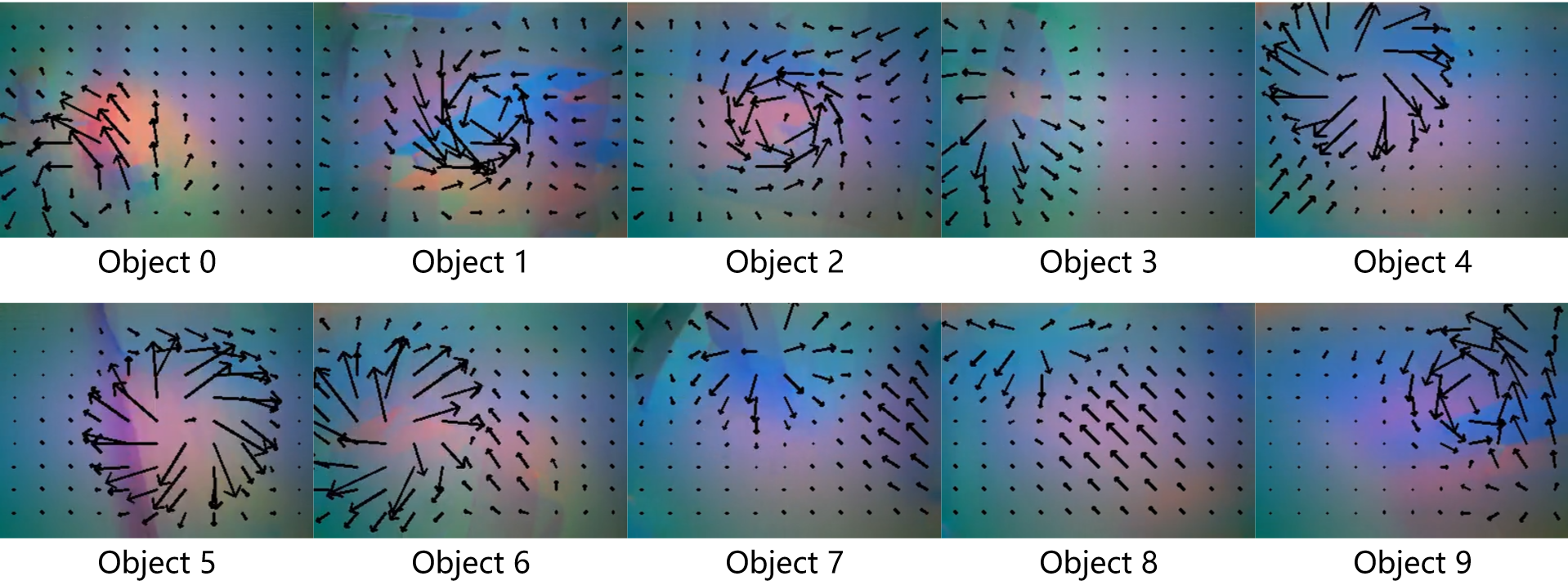}
    \caption{
    \textbf{Tactile observations across deformable assets.}
    Each panel shows a tactile RGB image with its marker-motion overlay
    at the moment of grasping, for the ten \textsc{Object-Soft} tasks.
    Objects that differ in geometry, compliance, and contact area
    produce distinct contact patches and shear fields under the same
    pick-and-place instruction. These are the cues available to a
    visuo-tactile policy and absent from external RGB views.
    }
    \label{fig:tactile_rgb_examples}
\end{figure*}

\paragraph{Visuo-Tactile Observations and Actions.}
The policy-visible data pair complementary visual and tactile views of
each manipulation. Third-person and wrist RGB describe the scene,
object, and target, while touch directly observes the contact interface
once the gripper engages the object. Each finger records tactile RGB and
marker-motion fields that capture contact appearance and local shear.
The tactile channels are rendered by
TacEx~\cite{nguyen2024tacex}
for simulated GelSight Mini sensors~\cite{yuan2017gelsight}, using
Taxim~\cite{si2022taxim} for optical contact appearance and
FOTS~\cite{zhao2024fots} for marker motion.
Figure~\ref{fig:tactile_rgb_examples} shows the resulting observations
across deformable assets: geometry, compliance, and contact area differ
per object and produce visibly different contact patches and shear
patterns, which are absent from external RGB views.

The remaining policy-visible channels provide robot state and task
specification. They include the end-effector pose, arm joint states,
current gripper width, the natural-language instruction, and the
executed arm and gripper actions. Each gripper action is stored both as
a binary open--close command and as a continuous closure target. This
dual encoding allows observation modality and
control granularity to be crossed without recollecting demonstrations.
Formally, the policy-visible observation is
\begin{equation}
o_t = \{I_t^{\mathrm{third}}, I_t^{\mathrm{wrist}},
\mathcal{T}_t, p_t, \ell\},
\end{equation}
and the recorded action is
\begin{equation}
a_t = (\mathbf{x}_t^{ee}, \boldsymbol{\theta}_t^{ee}, g_t),
\end{equation}
where $\mathcal{T}_t$ denotes the dual-finger tactile images and marker
fields, $p_t$ the proprioceptive state, $\ell$ the instruction,
$\mathbf{x}_t^{ee}$ and $\boldsymbol{\theta}_t^{ee}$ the absolute
end-effector position and axis-angle orientation, and $g_t$ the gripper
command in either encoding.

\paragraph{Evaluator-Only Physical States.}
The evaluator-only stream contains FEM nodal positions, object poses,
contact events, and drop events. It is written to disk alongside the
observations but is never exposed to a policy at training or evaluation
time; it is read only by the curation procedure of
Section~\ref{sec:curation} and by the evaluator of
Section~\ref{sec:benchmark}. Releasing this stream allows the
deformation criterion to be recomputed, audited, or redefined without
repeating data collection.

\subsection{Scene and Task Generation}
\label{sec:generation}

The four suites are diagnostic controls over object type (deformable or
rigid twin) and variation type (object or spatial). Object suites hold
the scene layout fixed and vary the manipulated asset, probing adaptation
to geometry and contact response. Spatial suites place two visually
identical instances in one scene and identify the target through the
language instruction, probing target grounding under a shared physical
layout. The underlying pick-and-place skill remains fixed, so differences
can be associated with object type or variation axis rather than a
different manipulation skill.

Within each task, episodes are generated by sampling from three
families of variation factors: visual factors such as lighting,
spatial factors such as object placement, target placement, and robot
initialization, and physical factors such as mass and stiffness. The
training ranges of these factors define the in-distribution support,
and held-out values strictly outside those ranges define the
out-of-distribution conditions of Section~\ref{sec:composition}. Only
one factor is shifted at a time, which is what makes the
out-of-distribution results attributable to a specific factor rather
than to a compound change in the scene.

\subsection{Demonstration Collection and Quality Control}
\label{sec:collection}
\label{sec:curation}

\paragraph{Collection and Synchronization.}
A scripted expert executes approach, grasp, lift, transport, and place
sequences. Each episode randomizes the object's initial pose, the robot
configuration, and the target location, and draws its grasp closure
from the calibrated gripper envelope $\mathcal{G}_o$ of
Section~\ref{sec:calibration}. Every trajectory is executed once under
physics and then rendered from its stored replay, so that visual,
tactile, proprioceptive, action, and FEM records share a single
timeline at $20$\,Hz and no stream is resampled or interpolated onto
another. Across the $2{,}000$ demonstrations on deformable assets, the
tolerance-normalized peak deformation $R_{\max}$ defined in
Section~\ref{sec:metrics} has a median of $0.433$ and a $95$th
percentile of $0.713$, and no demonstration exceeds its tolerance. The
tolerance violations reported for learned policies in
Section~\ref{sec:experiments} are therefore produced by those policies
and are not inherited from the supervision.

\paragraph{Quality Control.}
All collected episodes undergo automatic integrity screening and human
verification before inclusion, yielding the $4{,}000$ released
demonstrations.

\subsection{Physical Ground Truth and Per-Object Calibration for Evaluator}
\label{sec:calibration}

\paragraph{Deformation from Physical State.}
Each deformable asset is represented as a volumetric FEM mesh, so the
simulator tracks the position of every interior node throughout an
episode. Deformation is measured as the peak rigid-motion-removed RMS
displacement of these nodes, normalized by the initial bounding-box
diagonal so that assets of different sizes are comparable. Removing
rigid motion is what separates being carried from being squeezed, and
normalizing by object size is what makes a millimeter of compression
mean the same thing on a small roll and a large loaf. The measure is
computed from evaluator-only simulator states rather than inferred from
policy observations, and is negligible for the stiffness-raised rigid
twins.

\paragraph{Per-Object Calibration.}
Before policy training, a scripted grasp, lift, and hold protocol
sweeps the gripper closure from loose to tight. The first closure that
repeatedly holds the object without slip defines $g^{\min}_o$. The
deformation tolerance $\tau_o$ is the $90$th percentile of peak
displacements over stable grasps, and $g^{\max}_o$ is the tightest
calibrated closure that remains within this tolerance. Together, the
gripper envelope $\mathcal{G}_o=[g^{\min}_o,g^{\max}_o]$ and the
deformation tolerance $\tau_o$ form the per-object calibration. The
gripper envelope
$\mathcal{G}_o$ constrains expert collection, whereas $\tau_o$ supplies
the benchmark's deformation criterion. Because calibration precedes
policy training and uses only scripted physical probing, no evaluated
policy can influence the criterion against which it is scored. The
resulting tolerance is an operational benchmark criterion for
deformation compliance in simulation, not a material-failure or real-world damage
threshold.
Per-asset calibration parameters and the released-trace rescoring
procedure for alternative calibration percentiles are documented in
the appendix.

Together, the synchronized visual and tactile observations form the
policy-facing data interface, while evaluator-only FEM states and fixed
object-specific tolerances form the independent scoring interface used
by the closed-loop benchmark in Section~\ref{sec:benchmark}.

\section{The SoftVTBench Benchmark}
\label{sec:benchmark}

SoftVTBench evaluates whether a policy completes an instructed task while
maintaining acceptable physical interaction with the object. Policies
are executed in closed loop on the fixed evaluation sets of
Section~\ref{sec:composition}, using the same physics and observation
interface as data collection. The benchmark defines the policy
input--output interface, an evaluator-only physical state interface, the
Deformation-aware Success Rate (DSR), and paired ID/OOD protocols.

\subsection{Closed-Loop Evaluation Interface}
\label{sec:protocol}


At the $20$\,Hz control rate, the simulator executes actions from the policy's current predicted action chunk. At each policy query, the policy receives the policy-visible observation of Section~\ref{sec:streams}: third-person and wrist RGB, proprioception, the language instruction for language-conditioned policies, and, for visuo-tactile variants, dual-finger tactile images and marker fields. The policy predicts an action chunk consisting of absolute end-effector pose targets together with gripper commands in
the encoding on which it was trained, either binary or continuous. The predicted actions are then executed for a model-specific number of control steps before the policy replans. After each executed action, the simulator advances the robot, object, and contact states and returns the next observation, coupling perception and control through closed-loop physics rather than replaying a fixed trajectory. An
episode runs for at most $300$ control steps and terminates early on success or on a drop.

Throughout the rollout, the evaluator reads the evaluator-only stream of
FEM nodal positions, object poses, contact events, and drop events.
These states are never returned to the policy, so a policy must infer
contact conditions, incipient slip, and material compliance from its
visual, tactile, and proprioceptive inputs alone. All methods are
scored on identical episodes, initial states, and seeds.

\subsection{Deformation-aware Success Rate}
\label{sec:metrics}

\paragraph{Task Success Predicate.}
Task success requires the instructed object, rather than another
instance of the same type, to come to rest inside the designated target
region. This predicate is purely kinematic: deformation does not enter
it, which is exactly what makes it possible to measure how much a
task-success-only criterion misses.

\paragraph{Deformation Compliance and DSR.}
Let $T_i\in\{0,1\}$ indicate task success in episode $i$, and let
$D_{i,t}$ denote the rigid-motion-removed RMS displacement of its FEM
nodes at time $t$, normalized by the initial bounding-box diagonal.
Using the object-specific tolerance $\tau_{o_i}$ of
Section~\ref{sec:calibration}, we summarize episode-level deformation
using the normalized trace $R_t^{(i)}=D_{i,t}/\tau_{o_i}$ and its peak,
\begin{equation}
R_{\max}^{(i)}=\max_t R_t^{(i)}
=\max_t \frac{D_{i,t}}{\tau_{o_i}},
\end{equation}
so that $R_{\max}^{(i)}>1$ means the episode exceeded its calibrated
deformation tolerance, whether or not it recovered afterwards.

DSR is the primary benchmark metric. It is the fraction of evaluation
episodes that both satisfy the task-success predicate and remain within
the calibrated deformation tolerance. We also report Task Success Rate
(TSR), the fraction with $T_i=1$, as a diagnostic reference:
\begin{equation}
\mathrm{TSR}=\frac{1}{N}\sum_{i=1}^{N} T_i,
\qquad
\mathrm{DSR}=\frac{1}{N}\sum_{i=1}^{N}
T_i\cdot\mathbf{1}\!\left[R_{\max}^{(i)}\leq 1\right].
\end{equation}
Because task success is part of DSR, a policy that avoids contact and
never completes the task receives no credit. The difference
$\mathrm{TSR}-\mathrm{DSR}$ is exactly the
fraction of episodes that succeed while exceeding the deformation
tolerance, and $(\mathrm{TSR}-\mathrm{DSR})/\mathrm{TSR}$ is their
share among a policy's own successes. For the rigid suites, deformation
is negligible by construction and DSR coincides with TSR.

\subsection{In-Distribution and Out-of-Distribution Protocols}
\label{sec:protocols}

Each configuration is scored in two regimes. In-distribution evaluation
uses $500$ held-out episodes per suite and configuration, $50$ per
task, with all variation factors inside the training support.
Out-of-distribution evaluation applies the nine held-out conditions of
Section~\ref{sec:composition} to the two deformable suites, covering
three levels each of illumination, object mass, and stiffness with only
one parameter shifted at a time. Each condition reuses the task,
initial state, and seed of its in-distribution reference, so the
reported change is attributable to the shifted parameter. The nine
conditions together contribute $900$ episodes per suite and
configuration, and the reported out-of-distribution score is pooled
over them. Every $\Delta$ reported below is therefore measured against the
in-distribution entry for the same model, input, and suite.

Section~\ref{sec:experiments} instantiates this protocol with three
policy families and paired vision-only and visuo-tactile variants.

\section{Experiments}
\label{sec:experiments}

The experiments are organized around what the dataset and benchmark are
built to answer. We ask, in turn, whether process-level scoring exposes
physical-interaction failures that success-only evaluation accepts
(Section~\ref{sec:results}); how task completion and interaction
quality differ across policy families (Section~\ref{sec:family}); what
deformability costs relative to a matched rigid twin
(Section~\ref{sec:rigid}); how much of an apparent tactile gain is
attributable to sensing and how much to gripper control granularity
(Section~\ref{sec:ablation}); and when touch helps, across task
structures and under distribution shift (Section~\ref{sec:ood}).
Section~\ref{sec:validity} then examines whether the deformation
criterion itself carries information that completion does not.

\subsection{Does Process-Level Scoring Reveal What Success-Only
Evaluation Misses?}
\label{sec:results}

\begin{table*}[t]
\centering
\caption{
\textbf{In-distribution Task Success Rate and Deformation-aware Success
Rate on the deformable suites} (\%).
DSR is lower than TSR in every one of the twelve configurations; the
difference is the fraction of rollouts that reach the target while
leaving the calibrated interaction safety zone.
}
\label{tab:id_results}
\renewcommand{\arraystretch}{1.15}
\setlength{\tabcolsep}{8pt}
\small
\begin{tabular}{@{}ll rr rr@{}}
\toprule
& & \multicolumn{2}{c}{\textsc{Object-Soft}}
& \multicolumn{2}{c}{\textsc{Spatial-Soft}} \\
\cmidrule(lr){3-4}\cmidrule(l){5-6}
\textbf{Model} & \textbf{Input}
& \textbf{TSR} & \textbf{DSR}
& \textbf{TSR} & \textbf{DSR} \\
\midrule
\multirow{2}{*}{Diffusion Policy}
& VO-C & 37.4 & 33.6 & 15.6 & 13.4 \\
& VT-C & 40.0 & 30.4 & 33.0 & 25.0 \\
\cmidrule(lr){1-6}
\multirow{2}{*}{$\pi_{0.5}$}
& VO-C & 41.6 & 38.4 & 26.0 & 22.6 \\
& VT-C & 41.4 & 35.0 & 27.6 & 22.0 \\
\cmidrule(lr){1-6}
\multirow{2}{*}{FastWAM}
& VO-C & \textbf{62.0} & \textbf{58.0} & 37.0 & 36.6 \\
& VT-C & 57.6 & 54.4 & \textbf{56.4} & \textbf{56.0} \\
\bottomrule
\end{tabular}
\normalsize
\end{table*}

\noindent\textbf{Finding 1: Completion-only evaluation can misrank
policies because deformation violations are policy dependent.}
Table~\ref{tab:id_results} shows that the omission is consequential
rather than merely semantic. The gap is
non-zero in all twelve configurations, and for Diffusion Policy VT-C it
reaches $9.6$ percentage points on \textsc{Object-Soft} and $8.0$ on
\textsc{Spatial-Soft}, corresponding to $24\%$ of that configuration's
successful rollouts on each suite. Since TSR and DSR are computed on
the same rollouts, this is an exact count of $48$ and $40$ episodes
respectively, not a difference between two noisy estimates. These are
the episodes a completion-only protocol accepts and SoftVTBench
identifies as leaving the interaction safety zone. 
Because the tolerance is calibrated per object before policy training
and fixed across evaluated policies, differences in the TSR--DSR gap
cannot be explained by policy-specific or post-hoc adjustment of the
deformation criterion.

Figure~\ref{fig:qualitative} makes the distinction concrete. The upper
rows show rollouts that complete the task while remaining inside the
zone, with $R_{\max}$ between $0.28$ and $0.74$, peaking at the moment
of grasp and settling afterwards. The lower rows show rollouts that a
completion-only protocol scores identically, but whose $R_{\max}$ rises
above $1$ during the grasp and remains elevated through transport,
before a placement that is itself executed correctly. The third-person
and wrist views of the two cases are hard to tell apart; the marker
fields and the deformation trace are not.

\subsection{How Do Policy Families Differ in Completion and
Interaction?}
\label{sec:family}

The hidden violation rate varies substantially across policy families.
Across the configurations of Table~\ref{tab:id_results}, the share of
successful rollouts falling outside the safety zone spans $10$--$24\%$
for Diffusion Policy and $8$--$20\%$ for $\pi_{0.5}$, but only
$0.7$--$6.5\%$ for FastWAM. On both spatial configurations, FastWAM
keeps TSR and DSR within $0.4$ percentage points, that is two episodes
out of $500$, while attaining the strongest spatial results in the
table. Completion and interaction quality are
therefore not two names for the same ranking: SoftVTBench separates a
policy that succeeds by regulating contact from one that succeeds
despite mishandling the object, a distinction terminal success cannot
express. It also shows that close alignment between the two criteria is
attainable rather than precluded by the domain.

\subsection{What Does Deformability Cost Relative to a Rigid Twin?}
\label{sec:rigid}

\begin{table*}[t]
\centering
\caption{
\textbf{Task success on deformable assets and their rigid twins}
(TSR, \%).
Twins match geometry, appearance, and mass under the same layouts and
instructions, with negligible deformation, so a rigid--deformable
difference isolates the effect of deformability. Because deformation is
negligible on the twins, DSR coincides with TSR there and only TSR is
reported.
}
\label{tab:rigid_twins}
\renewcommand{\arraystretch}{1.15}
\setlength{\tabcolsep}{8pt}
\small
\begin{tabular}{@{}ll cc cc@{}}
\toprule
& & \multicolumn{2}{c}{Object variation}
& \multicolumn{2}{c}{Spatial variation} \\
\cmidrule(lr){3-4}\cmidrule(l){5-6}
\textbf{Model} & \textbf{Input}
& \textbf{Rigid} & \textbf{Soft}
& \textbf{Rigid} & \textbf{Soft} \\
\midrule
\multirow{2}{*}{Diffusion Policy}
& VO-C & 40.0 & 37.4 & 14.0 & 15.6 \\
& VT-C & 35.0 & 40.0 & 11.0 & 33.0 \\
\cmidrule(lr){1-6}
\multirow{2}{*}{$\pi_{0.5}$}
& VO-C & 60.0 & 41.6 & \textbf{50.4} & 26.0 \\
& VT-C & 59.6 & 41.4 & \textbf{54.0} & 27.6 \\
\cmidrule(lr){1-6}
\multirow{2}{*}{FastWAM}
& VO-C & \textbf{64.0} & \textbf{62.0} & 25.0 & \textbf{37.0} \\
& VT-C & \textbf{61.6} & \textbf{57.6} & 30.0 & \textbf{56.4} \\
\bottomrule
\end{tabular}
\normalsize
\end{table*}

\noindent\textbf{Finding 2: The apparent cost of deformability is not
an intrinsic property of soft objects, but depends on policy and task
structure.}
The matched VO-C pairs in Table~\ref{tab:rigid_twins} isolate the
effect of deformability under otherwise identical task conditions, and
two of the four resolved effects are large enough to interpret. On
object variation, $\pi_{0.5}$ loses $18.4$ percentage points from rigid
to soft, whereas Diffusion Policy and FastWAM lose $2.6$ and $2.0$
points, both inside the resolution of the protocol and therefore not
distinguishable from no cost at all. On spatial variation the sign
reverses: FastWAM scores $12.0$ points \emph{higher} on the deformable
assets, and Diffusion Policy $1.6$ points higher, again within noise.
A negative cost of deformability suggests that factors other than contact physics, such as target grounding or spatial generalization, may dominate task difficulty in that configuration. This is precisely the type of confounding that a soft-only benchmark cannot disentangle.
The rigid twins therefore provide a necessary control for separating the effect of deformability from difficulty intrinsic to the task structure.

Two caveats bound this comparison. The rigid visuo-tactile rows of
Table~\ref{tab:rigid_twins} decode binary gripper execution from
continuous-policy checkpoints rather than being trained natively, and
are therefore corroborating rather than primary evidence. Separately,
Diffusion Policy is not language-conditioned, so on the spatial suites, where the
target instance is designated only by the instruction, it cannot in
principle distinguish the two identical instances; 
its spatial performance is confounded by the absence of language conditioning.

\subsection{How Much of the Tactile Benefit Is Sensing, and How Much
Is Control Granularity?}
\label{sec:ablation}

A comparison between a binary vision-only policy and a continuous
visuo-tactile one confounds observation modality with control
granularity. Because SoftVTBench stores both gripper-action encodings
for every demonstration (Section~\ref{sec:streams}), the two factors
can be crossed without recollecting data,.

\begin{table*}[t]
\centering
\caption{
\textbf{Matched sensing--control ablation for $\pi_{0.5}$} (\%).
VO/VT denote vision-only and visuo-tactile inputs; B/C denote binary
and continuous gripper control. Crossing the two factors separates a
gain due to touch from a gain due to finer actuation.
}
\label{tab:pi05_ablation}
\renewcommand{\arraystretch}{1.15}
\setlength{\tabcolsep}{10pt}
\small
\begin{tabular}{@{}l rr rr@{}}
\toprule
& \multicolumn{2}{c}{\textsc{Object-Soft}}
& \multicolumn{2}{c}{\textsc{Spatial-Soft}} \\
\cmidrule(lr){2-3}\cmidrule(l){4-5}
\textbf{Configuration} & \textbf{TSR} & \textbf{DSR}
& \textbf{TSR} & \textbf{DSR} \\
\midrule
VO-B & 30.2 & 27.2 & \textbf{34.2} & 20.0 \\
VO-C & \textbf{41.6} & \textbf{38.4} & 26.0 & \textbf{22.6} \\
VT-B & 41.0 & 28.0 & 30.0 & 21.4 \\
VT-C & 41.4 & 35.0 & 27.6 & 22.0 \\
\bottomrule
\end{tabular}
\normalsize
\end{table*}

\noindent\textbf{Finding 3: Tactile gains are not identifiable unless
sensing modality and control granularity are matched.}
Starting from VO-B on \textsc{Object-Soft}, continuous control alone
(VO-C) raises TSR by $11.4$ percentage points, while tactile input
alone (VT-B) raises it by $10.8$ points. Combining both (VT-C) yields
$41.4\%$, no improvement over continuous control by itself. A study
that compared VO-B against VT-C would therefore credit touch with a
gain that finer actuation reproduces on its own. Storing both encodings
is what allows the benchmark to make this distinction.

The same ablation also shows why completion alone is insufficient for
this attribution. VO-C and VT-B differ by only $0.6$ percentage points in
\textsc{Object-Soft} TSR, yet VO-C is $10.4$ points higher in DSR
($38.4\%$ vs.\ $28.0\%$): $7.7\%$ of VO-C successes leave the safety
zone, against $31.7\%$ for VT-B. Continuous control raises DSR in all
four within-modality binary-to-continuous comparisons across the two
suites. The effect is bounded to $\pi_{0.5}$ and depends on task
structure, since on \textsc{Spatial-Soft} both upgrades reduce TSR
relative to VO-B, consistent with target grounding rather than contact
regulation being the binding constraint there. The ablation
nevertheless establishes that action spaces must be matched before any
performance difference is attributed to tactile sensing.

\subsection{When Does Touch Help, and Under What Shift?}
\label{sec:ood}

\begin{table*}[t]
\centering
\caption{
\textbf{Out-of-distribution Task Success Rate and Deformation-aware
Success Rate on the deformable suites} (\%), pooled over the nine
held-out conditions of Section~\ref{sec:composition}.
$\Delta$ is the change relative to the corresponding in-distribution
entry for the same model, input, and suite in
Table~\ref{tab:id_results}.
}
\label{tab:ood_results}
\renewcommand{\arraystretch}{1.1}
\setlength{\tabcolsep}{6pt}
\small
\begin{tabular}{@{}ll rr rr@{}}
\toprule
& & \multicolumn{2}{c}{\textsc{Object-Soft}}
& \multicolumn{2}{c}{\textsc{Spatial-Soft}} \\
\cmidrule(lr){3-4}\cmidrule(l){5-6}
\textbf{Model} & \textbf{Input}
& \makecell[c]{TSR $\uparrow$\\($\Delta$ vs.\ ID)}
& \makecell[c]{DSR $\uparrow$\\($\Delta$ vs.\ ID)}
& \makecell[c]{TSR $\uparrow$\\($\Delta$ vs.\ ID)}
& \makecell[c]{DSR $\uparrow$\\($\Delta$ vs.\ ID)} \\
\midrule
\multirow{2}{*}{Diffusion Policy}
& VO-C & 29.2 ($-$8.2) & \textbf{26.6 ($-$7.0)}
& 11.0 ($-$4.6) & 8.8 ($-$4.6) \\
& VT-C & \textbf{31.2 ($-$8.8)} & 25.0 ($-$5.4)
& \textbf{25.2 ($-$7.8)} & \textbf{17.8 ($-$7.2)} \\
\cmidrule(lr){1-6}
\multirow{2}{*}{$\pi_{0.5}$}
& VO-C & 35.8 ($-$5.8) & 33.2 ($-$5.2)
& 24.4 ($-$1.6) & 19.4 ($-$3.2) \\
& VT-C & \textbf{41.0 ($-$0.4)} & \textbf{34.2 ($-$0.8)}
& \textbf{28.4 ($+$0.8)} & \textbf{23.2 ($+$1.2)} \\
\cmidrule(lr){1-6}
\multirow{2}{*}{FastWAM}
& VO-C & 54.4 ($-$7.6) & 53.8 ($-$4.2)
& 27.8 ($-$9.2) & 27.2 ($-$9.4) \\
& VT-C & \textbf{55.8 ($-$1.8)} & \textbf{55.8 ($+$1.4)}
& \textbf{39.4 ($-$17.0)} & \textbf{38.8 ($-$17.2)} \\
\bottomrule
\end{tabular}
\normalsize
\end{table*}

\begin{figure*}[t]
    \centering
    \includegraphics[width=1\linewidth]{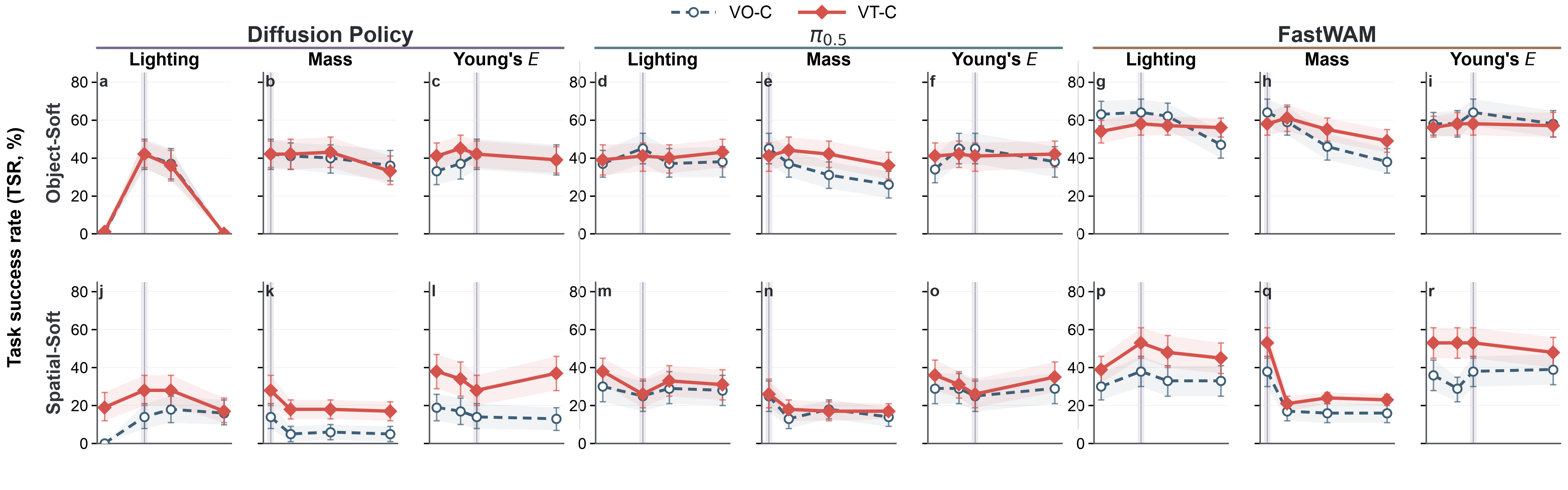}
    \caption{
    \textbf{Task Success Rate under out-of-distribution shifts,
    resolved by factor.}
    Rows correspond to the \textsc{Object-Soft} and
    \textsc{Spatial-Soft} suites; column groups correspond to Diffusion
    Policy, $\pi_{0.5}$, and FastWAM, each resolved into illumination,
    object-mass, and Young's-modulus shifts.
    }
    \label{fig:ood_trends}
\end{figure*}

\noindent\textbf{Finding 4: Tactile sensing is more consistently
associated with robustness under distribution shift than with peak
in-distribution performance, but completion gains need not improve
deformation compliance.}
Table~\ref{tab:ood_results} pools the nine held-out conditions. VT-C
exceeds VO-C in TSR in all six policy--suite comparisons and in DSR in
five, with Diffusion Policy on \textsc{Object-Soft} the sole exception.
Several of the individual margins are small, but the consistency of the
direction is itself the result: under the one-sided exact sign test, six of six agreeing
comparisons has $p=0.016$, and five of six has $p=0.11$. In
distribution the same comparison is split, with VT-C ahead of VO-C on
TSR in four of six and behind in two
(Table~\ref{tab:id_results}), which is consistent with chance-level directional variation at this sample size.
The value of contact sensing is therefore better read as robustness
than as peak in-distribution performance.

Figure~\ref{fig:ood_trends} resolves this by factor, and the pattern is organized more by suite than by factor. On \textsc{Spatial-Soft} (bottom row), the visuo-tactile curve generally sits at or above the vision-only curve across the three factors and policy families, which
contributes to the aggregate result above. On \textsc{Object-Soft} (top row), the two curves overlap more closely overall, although policy-dependent separations appear under several shifts, including
mass shifts for $\pi_{0.5}$ and FastWAM. Tactile cues become available after contact and may therefore provide information about contact dynamics induced by mass changes. In contrast, illumination perturbs the visual observations throughout the rollout, including the
pre-contact approach phase. This may help explain why Diffusion Policy degrades strongly at the illumination extremes on \textsc{Object-Soft} under both modalities, although the benchmark does not isolate the causal mechanism.


However, under shift, adding touch also widens the
$\mathrm{TSR}-\mathrm{DSR}$
gap for the two policies whose gap is large to begin with: for
Diffusion Policy it grows from $2.6$ to $6.2$ points on
\textsc{Object-Soft} and from $2.2$ to $7.4$ on \textsc{Spatial-Soft},
and for $\pi_{0.5}$ from $2.6$ to $6.8$ points on \textsc{Object-Soft}.
In the remaining three comparisons the gap changes by at most $0.6$
points and is unchanged at this resolution. FastWAM is the constructive
case: under both modalities its TSR and DSR stay within one percentage
point in every pooled comparison, so its robustness gain from touch
does not come at the cost of interaction quality. Improvements along
the completion axis and along the interaction axis can align, but they
do not do so automatically, and this is exactly the divergence a
completion-only protocol would record as uniform progress.

\subsection{Does the Deformation Criterion Add Information?}
\label{sec:validity}

As the primary criterion, DSR should carry information beyond
completion and should reflect policy behavior rather than an
unreachable tolerance.

\noindent\textbf{The criterion is discriminative.}
On \textsc{Object-Soft}, TSR ranks Diffusion Policy VT-C above VO-C
($40.0\%$ vs.\ $37.4\%$), whereas DSR reverses the ranking ($30.4\%$
vs.\ $33.6\%$). The matched ablation supplies a second case: VO-C and
VT-B differ by $0.6$ points in TSR but by $10.4$ points in DSR. DSR
therefore changes conclusions that completion alone would support,
rather than acting as a proxy for it.

\noindent\textbf{The criterion is attainable.}
Expert collection is constrained by the calibrated gripper envelopes, and empirically none of the accepted demonstrations on deformable assets exceeds the calibrated deformation tolerance. Moreover, the
observed deformation remains well below the threshold for most demonstrations: across the 2,000 demonstrations on deformable assets, the median $R_{\max}$ is $0.433$ and the 95th percentile is $0.713$ (Section~\ref{sec:collection}). These demonstrations successfully cover the benchmark tasks while remaining deformation-compliant, showing that satisfying the deformation criterion is compatible with
task completion. The violations recorded in all twelve learned-policy configurations are therefore not inherited from the supervision. Sensitivity to the calibration percentile is reported in the technical appendix.

\noindent\textbf{Alignment is achievable.}
FastWAM attains the highest TSR while leaving the smallest share of its
successes outside the safety zone in all four in-distribution
configurations ($0.7$--$6.5\%$), and keeps TSR and DSR within one
percentage point under both modalities in the pooled
out-of-distribution conditions. This is an existence proof that the
success--safety gap is a tractable policy and control problem rather
than a fixed cost of handling deformable objects.


\section{Conclusion}
\label{sec:conclu}

We presented SoftVTBench, a visuo-tactile dataset and closed-loop
benchmark for physical-interaction-aware deformable-object manipulation.
Its central data contribution is $4{,}000$ expert demonstrations across
$40$ pick-and-place tasks, pairing multi-view RGB with dual-finger
tactile RGB and marker motion and synchronizing them with
proprioception, language, actions, and evaluator-only FEM states.
Matched rigid twins, controlled variations, synchronized recording, and
auditable curation make the resource suitable for both policy learning
and diagnostic evaluation. Built on these data, the benchmark uses DSR
as its primary metric, counting a rollout as successful only when it
completes the task and satisfies an object-specific deformation
tolerance under closed-loop ID and OOD evaluation.

Experiments across three policy families demonstrate the benchmark's
necessity and diagnostic value. All twelve ID deformable-object
configurations include rollouts that achieve task success while exceeding
the calibrated deformation tolerance, accounting for $0.7$--$24\%$
of each configuration's successes. The matched controls further show
that physical-interaction performance depends on policy family and
deformability, that gripper control granularity can reproduce some
apparent tactile gains, and that visuo-tactile variants attain higher OOD
TSR in all six comparisons and higher OOD DSR in five of six, despite
mixed ID effects.
Across a policy trained from scratch, an adapted vision-language-action
model, and a world--action model, these mixed ID results show that
reliably fusing tactile cues remains an open modeling problem. FastWAM's
small TSR--DSR gap nevertheless shows that task success and deformation
compliance can align, rather than imposing an unavoidable tradeoff.


\clearpage
\bibliographystyle{bibstyle}
\bibliography{main}

\newpage
\setcounter{section}{0}
\section{Overview}

This supplement provides implementation details, per-object
calibration results, task-suite specifications, baseline training
settings, per-condition out-of-distribution results, additional
analyses, and the scope of tactile simulation.
Table~\ref{tab:navigation} summarizes the contents in section order.

\begin{table}[!h]
\centering
\small
\renewcommand{\arraystretch}{1.15}
\begin{tabular}{@{}p{0.34\linewidth}p{0.60\linewidth}@{}}
\toprule
\textbf{Supplementary section} & \textbf{Contents} \\
\midrule
Sec.~\ref{sec:implementation}: Simulation and implementation
details & Simulator, robot, sensing, FEM, and compute configuration. \\
Sec.~\ref{sec:assets}: Assets and interaction safety zones & Asset
inventory, rigid twins, per-object calibration, and the calibration
percentile. \\
Sec.~\ref{sec:tasks}: Task suite details & Suite design, task mapping,
and variation factors. \\
Sec.~\ref{sec:training}: Baseline training details & Training and
inference settings for all evaluated policy families. \\
Sec.~\ref{sec:ood_per_condition}: Per-condition results
& Out-of-distribution results per policy family and condition. \\
Sec.~\ref{sec:analyses}: Additional analyses & Qualitative cases and
stiffness-conditioned deformation statistics. \\
Sec.~\ref{sec:tactile}: Tactile simulation pipeline and scope & Tactile
rendering stages, recorded outputs, and validity boundary. \\
\bottomrule
\end{tabular}
\caption{Contents of the supplementary sections in document order.}
\label{tab:navigation}
\end{table}

\section{Simulation and Implementation Details}
\label{sec:implementation}

SoftVTBench is implemented in Isaac Sim~4.5.0 with Isaac Lab~0.41.3 and
the GPU-accelerated PhysX~5 pipeline. Physics runs at 60\,Hz with a
control decimation of 3, giving a 20\,Hz control and logging rate; all
visual, tactile, proprioceptive, action, and evaluator-only physical-state streams
are synchronized at this rate. Table~\ref{tab:implementation_details}
summarizes the robot, control, sensing, simulation, and hardware
configuration.

\begin{table}[!h]
\centering
\renewcommand{\arraystretch}{1.12}
\setlength{\tabcolsep}{4pt}
\small
\begin{tabular}{@{}p{0.30\linewidth}p{0.64\linewidth}@{}}
\toprule
\textbf{Item} & \textbf{Value} \\
\midrule
Simulator & Isaac Sim 4.5.0 / Isaac Lab 0.41.3, PhysX 5 GPU pipeline \\
Physics / control rate & 60\,Hz physics, decimation 3, 20\,Hz control \\
Robot & Franka arm with Panda parallel-jaw gripper \\
Controller & Task-space differential inverse kinematics \\
End-effector action & Absolute pose target: 3D position and 3D axis-angle orientation \\
Gripper action & Normalized closure command; continuous and binary encodings \\
Finger friction & Static $\mu_s=1.5$, dynamic $\mu_d=1.2$, max combine mode \\
Camera views & Third-person $1024{\times}1024$ and wrist $512{\times}512$, resized to $224{\times}224$ \\
Tactile sensor & GelSight Mini via TacEx; Taxim optics and FOTS markers \\
Tactile streams & Tactile RGB and $11{\times}9$ marker-motion field, $320{\times}240$ \\
FEM model & PhysX soft body with corotational linear elasticity \\
FEM solver & Hex resolution 6, 64 position iterations, damping 2.5 \\
Collection / evaluation & $4{\times}$ NVIDIA L20 GPU \\
Training & NVIDIA A100-80GB; per-family counts in
Table~\ref{tab:baseline_training_details} \\
\bottomrule
\end{tabular}
\caption{Simulation and sensing configuration.}
\label{tab:implementation_details}
\end{table}

\section{Assets and Per-Object Interaction Safety Zones}
\label{sec:assets}

\paragraph{Deformable assets.}
The dataset contains ten volumetric deformable assets: six
bakery-style meshes and four procedurally generated geometric
primitives. The surface meshes are converted into simulation-ready volumetric
meshes and the deformable instances are simulated as PhysX GPU FEM
soft bodies with corotational linear elasticity. Per-asset density,
friction, elasticity, and damping are authored in the simulation
assets and are never provided to the policy. Before each episode the
object is settled on the support surface; the expert and the evaluator
both use this settled state rather than the nominal spawn pose, so that
pre-grasp drift does not contaminate either data collection or scoring.

\paragraph{Rigid twins.}
Each rigid twin shares the mesh, texture, and mass of its deformable
counterpart, with stiffness raised so that deformation is negligible
under any achievable grasp. Twins populate the \textsc{Object-Rigid}
and \textsc{Spatial-Rigid} suites under the same layouts, instructions,
and recording stack as their soft counterparts.

\paragraph{Per-asset interaction safety zones.}
Table~\ref{tab:asset_zones} reports, for each of the ten deformable
assets, the calibrated gripper envelope
$\mathcal{G}_{o}=[g^{\min}_{o},g^{\max}_{o}]$, its physical aperture
span, and the normalized deformation tolerance $\tau_{o}$. The
envelope endpoints are the per-object compression ratios recorded in
the locked collection cards (larger values denote tighter grasps).
The millimeter span is computed directly as the difference between the
card's loose and tight measured jaw apertures; it is not inferred from
a universal linear command-to-width conversion.

\begin{table}[!h]
\centering
\renewcommand{\arraystretch}{1.1}
\setlength{\tabcolsep}{4pt}
\small
\begin{tabular}{@{}lrrrr@{}}
\toprule
\textbf{Asset} & $g^{\min}_{o}$ & $g^{\max}_{o}$ &
\textbf{Span (mm)} & $\tau_{o}$/diag.\ (\%) \\
\midrule
\texttt{soft\_pastry001}       & 0.4000 & 0.8000 & 26.21 & 7.6 \\
\texttt{soft\_pastry002}       & 0.6547 & 0.6900 &  2.56 & 9.3 \\
\texttt{soft\_pastry003}       & 0.4400 & 0.8000 & 13.26 & 8.6 \\
\texttt{soft\_pastry005}       & 0.4700 & 0.5800 &  5.87 & 10.7 \\
\texttt{soft\_pastry010}       & 0.4200 & 0.6600 & 24.82 & 8.3 \\
\texttt{soft\_pastry011}       & 0.4400 & 0.5800 & 10.63 & 7.1 \\
\midrule
\texttt{soft\_stw\_cube\_hq}    & 0.3200 & 0.6200 & 19.43 & 11.2 \\
\texttt{soft\_stw\_cuboid\_hq}  & 0.3800    & 0.4400  & 10.95 & 9.8 \\
\texttt{soft\_stw\_cylinder\_hq}& 0.2900 & 0.6600 & 21.95 & 9.8 \\
\texttt{soft\_stw\_sphere\_hq}  & 0.3700 & 0.6300 & 20.77 & 9.7 \\
\bottomrule
\end{tabular}
\caption{Interaction safety zones of the ten deformable assets.
$\tau_{o}$ is the object-specific 90th-percentile
threshold used by the released evaluator, expressed as a percentage of
the reference bounding-box diagonal.}
\label{tab:asset_zones}
\end{table}

\paragraph{Calibration percentile.}
All thresholds in Table~\ref{tab:asset_zones} use the 90th percentile
of peak displacement over the stable scripted grasps of the
calibration sweep. The percentile is a single global constant: it is
fixed once, before any policy is trained, and the same value is
applied to every asset, so it cannot be tuned per object or per
method. A lower percentile tightens every threshold uniformly and a
higher one loosens them, which rescales $R_{\max}$ by a per-object
constant and can only change a policy's score through episodes whose peak deformation lies between the two thresholds. Because the released evaluation records store the per-episode deformation trace $R_t$ rather than only the binary outcome, scoring
at another percentile is a re-aggregation of the released records and requires no new rollouts; the accompanying code package includes the rescoring entry point for
this purpose.

\section{Task Suite Details}
\label{sec:tasks}

The four suites form a matched $2\times2$ design over object type
(deformable or rigid twin) and variation axis (object identity or
spatial layout). Each suite contains 10 tasks with 100 expert
demonstrations per task, for 1{,}000 demonstrations per suite and
4{,}000 in total. All suites share the same robot embodiment, camera
views, tactile interfaces, rollout API, and evaluation protocol.

\textsc{Object-Soft} fixes the scene layout and varies the manipulated
object across tasks, so that the policy must adapt its grasp to each
object's geometry, compliance, and contact surface.
\textsc{Spatial-Soft} places two visually identical instances of the
same object in each scene and identifies the target through a spatial
referring expression in the language instruction, so that the scene
alone does not determine the referent. The success predicate checks
the identity of the transported instance and requires that the
instructed one, not merely some instance, reaches the target.
\textsc{Object-Rigid} and \textsc{Spatial-Rigid} replicate both
structures with the corresponding rigid twins.

\paragraph{Task list.}
Table~\ref{tab:task_list} lists the tasks of the two deformable
suites; the rigid suites replicate them with the corresponding twins
under identical layouts and instructions.

\begin{table}[!h]
\centering
\renewcommand{\arraystretch}{1.1}
\setlength{\tabcolsep}{4pt}
\small
\footnotesize
\begin{tabular}{@{}llp{0.55\linewidth}l@{}}
\toprule
\textbf{Suite} & \textbf{Task} & \textbf{Instruction} & \textbf{Asset} \\
\midrule
\multirow{10}{*}{\textsc{Object-Soft}}
& 0 & Pick up the white swirled pastry and place it in the basket. & \texttt{soft\_pastry001} \\
& 1 & Pick up the panda-face pastry and place it in the basket. & \texttt{soft\_pastry002} \\
& 2 & Pick up the small chocolate rectangular pastry and place it in the basket. & \texttt{soft\_pastry003} \\
& 3 & Pick up the soft cream-colored cube and place it in the basket. & \texttt{soft\_stw\_cube\_hq} \\
& 4 & Pick up the yellow square layered pastry and place it in the basket. & \texttt{soft\_pastry005} \\
& 5 & Pick up the soft tan cylinder and place it in the basket. & \texttt{soft\_stw\_cylinder\_hq} \\
& 6 & Pick up the golden pastry and place it in the basket. & \texttt{soft\_pastry011} \\
& 7 & Pick up the golden knotted pastry and place it in the basket. & \texttt{soft\_stw\_cuboid\_hq} \\
& 8 & Pick up the soft red ball and place it in the basket. & \texttt{soft\_stw\_sphere\_hq} \\
& 9 & Pick up the orange pumpkin-shaped pastry and place it in the basket. & \texttt{soft\_pastry010} \\
\midrule
\multirow{10}{*}{\textsc{Spatial-Soft}}
& 0 & Pick up the left white swirled pastry and place it on the plate. & \texttt{soft\_pastry001} \\
& 1 & Pick up the right panda-face pastry and place it on the plate. & \texttt{soft\_pastry002} \\
& 2 & Pick up the left small chocolate rectangular pastry and place it on the plate. & \texttt{soft\_pastry003} \\
& 3 & Pick up the left soft cream-colored cube and place it on the plate. & \texttt{soft\_stw\_cube\_hq} \\
& 4 & Pick up the left yellow square layered pastry and place it on the plate. & \texttt{soft\_pastry005} \\
& 5 & Pick up the left soft tan cylinder and place it on the plate. & \texttt{soft\_stw\_cylinder\_hq} \\
& 6 & Pick up the right golden pastry and place it on the plate. & \texttt{soft\_pastry011} \\
& 7 & Pick up the left soft cream-colored rectangular block and place it on the plate. & \texttt{soft\_stw\_cuboid\_hq} \\
& 8 & Pick up the left soft red ball and place it on the plate. & \texttt{soft\_stw\_sphere\_hq} \\
& 9 & Pick up the left orange pumpkin-shaped pastry and place it on the plate. & \texttt{soft\_pastry010} \\
\bottomrule
\end{tabular}
\caption{Task specifications for the deformable suites. Task identifiers
follow the released suite configuration and are zero-indexed;
instructions are reproduced verbatim from that configuration, and
object descriptions are therefore not normalized across suites.}
\label{tab:task_list}
\end{table}

\paragraph{Variation factors.}
Table~\ref{tab:variation_factors} lists the training ranges of the
variation families and the held-out values used for out-of-distribution
evaluation. Held-out values lie strictly outside the training ranges,
and each out-of-distribution condition shifts a single factor while
holding the others at their nominal values.

\begin{table}[!h]
\centering
\renewcommand{\arraystretch}{1.1}
\setlength{\tabcolsep}{4pt}
\small
\begin{tabular}{@{}llll@{}}
\toprule
\textbf{Family} & \textbf{Factor} & \textbf{Training range} &
\textbf{Held-out values} \\
\midrule
Visual & Dome-light intensity & 135 (fixed nominal) & 67.5, 180, 270 \\
Spatial & Object placement & Per-task recorded support & --- \\
        & Target placement & Per-task recorded support & --- \\
        & Robot initialization & Per-task recorded support & --- \\
Physical & Mass scale & $\times$1.0 (nominal) & $\times$1.25, $\times$1.75, $\times$2.5 \\
         & Young's modulus scale & $\times$1.0 (nominal) & $\times$0.5, $\times$0.8, $\times$2.0 \\
\bottomrule
\end{tabular}
\caption{Variation factors: training ranges and held-out
out-of-distribution values. The nine out-of-distribution conditions
are the three held-out levels of dome-light intensity, mass, and
Young's modulus.}
\label{tab:variation_factors}
\end{table}

\section{Baseline Training Details}
\label{sec:training}

We evaluate eight policy configurations. Six are primary
continuous-control baselines: the vision-only (VO-C) and
visuo-tactile (VT-C) variants of Diffusion Policy, $\pi_{0.5}$, and
FastWAM. The remaining two, VO-B and VT-B, are binary-control
ablations of $\pi_{0.5}$. These ablations differ from their continuous
counterparts only in the gripper-action encoding used during training
and inference; all other settings remain unchanged.

Unless stated otherwise, each method follows its official backbone,
optimizer family, learning-rate schedule, and model-specific inference
procedure. Benchmark-specific adaptations are restricted to the
observation interfaces---including the tactile encoders and
marker-motion adapters described below---the 7D action interface, and
micro-batch sizes selected to fit GPU memory. The added tactile
streams increase activation memory, so the visuo-tactile variants of
Diffusion Policy and FastWAM use a smaller micro-batch than their
vision-only counterparts; since no gradient accumulation is applied in
either configuration, their effective global batch size differs
accordingly (Table~\ref{tab:vo_vt_variant_details}). The two
$\pi_{0.5}$ variants are matched at a global batch size of 256.
Comparisons between vision-only and visuo-tactile variants therefore
reflect this optimization difference in addition to the observation
interface. Each reported
configuration corresponds to a single training run and uses the final
checkpoint from the fixed schedule listed in
Table~\ref{tab:baseline_training_details}.
We apply neither validation-based checkpoint selection nor a shared
cross-family hyperparameter sweep. Consequently, differences across
policy families may reflect both their architectures and their
prescribed training recipes.

\subsection{Diffusion Policy (DP)}

We retain the image-conditioned diffusion U-Net architecture and DDPM
parameterization of Diffusion Policy. The vision-only variant receives
two consecutive observations, each comprising a third-person RGB
image, a wrist RGB image, and the 7D robot state. Their representations
form the global conditioning vector of the U-Net, which denoises a
16-step chunk of 7D actions. DP is not language-conditioned.

The visuo-tactile variant preserves the same policy and action
interface while adding left- and right-finger tactile RGB as two image
streams. A ResNet-18 image encoder processes each tactile stream, and
the resulting representations are fused with the visual features. The
two-frame marker-motion representation is additionally concatenated
with the robot-state conditioning vector, providing a 792D
marker-motion input alongside the 7D robot state. Apart from this
observation interface, the VO and VT variants share the same network
design and normalization procedure. Their common optimization and
inference settings are reported in
Table~\ref{tab:baseline_training_details}.

\subsection{\boldmath$\pi_{0.5}$ Policy}

We initialize both modality variants from the pretrained $\pi_{0.5}$
model. The vision-only (VO) variant receives third-person RGB, wrist
RGB, the 7D robot state, and the language instruction. Except for the
task-specific observation and action transformations, VO follows the
official LoRA fine-tuning architecture and optimization procedure.
LoRA adapters are applied to both the PaliGemma backbone and the action
expert, while the pretrained backbone parameters remain frozen.

The visuo-tactile (VT) variant retains all VO inputs and additionally
receives tactile RGB and marker-motion observations from both fingers.
At each policy query, we collect the past eight tactile RGB frames from
each finger, yielding 16 frames in total, and tile them temporally into
a single $4{\times}4$ mosaic. The shared visual encoder processes this
mosaic in the same manner as the external RGB observations. This design
represents tactile appearance and contact deformation in the visual
token space without introducing a separate tactile image backbone.
Marker motion is encoded as low-dimensional contact features: the
current marker observation is injected into the policy prefix, whereas
a two-frame marker-motion history is supplied to the action suffix to
condition action generation on recent contact dynamics. The additional
tactile projection layers are trained jointly with the LoRA adapters.

Both variants predict a 50-step chunk of 7D actions comprising 3D
position, 3D axis-angle orientation, and a 1D gripper command. During
closed-loop execution, the policy executes 10 predicted actions before
replanning. Unless otherwise specified, each policy is LoRA-fine-tuned
for 7k steps on eight NVIDIA A100-80GB GPUs with a global batch size of
256. The complete training and inference configuration is reported in
Table~\ref{tab:baseline_training_details}.

\subsection{FastWAM}

We retain the official FastWAM world--action model architecture and
flow-matching formulation. In the vision-only setting, the RGB video
expert receives the third-person and wrist RGB observations, while the
action expert predicts the 7D action sequence. Cached tokens from the
frozen T5 encoder represent the language instruction, and a learnable
projection of the 7D robot state is appended to the language context
through cross-attention.

The visuo-tactile variant extends the original two-expert
mixture-of-transformers (MoT) with a third tactile DiT expert. At each
observation time, tactile input is constructed as a $2{\times}2$ mosaic
containing the left and right tactile images from the current and
preceding frames. The frozen Wan VAE encodes this mosaic, which is then
processed by the tactile expert. The tactile expert reuses the
ActionDiT-style pretrained backbone, while its tactile input and output
layers are learned for the tactile latent space. For marker motion, we
use the current frame together with the preceding two frames. After
normalization, a lightweight MLP maps the three marker-motion vectors
to temporally indexed context tokens. These tokens are appended only to
the text--proprioceptive context of the tactile and action experts,
leaving the RGB expert unchanged.

We further use an anchor-only tri-branch attention mask: the action
expert can attend to the observed first-frame RGB and tactile tokens but
never to future tactile tokens. Consequently, inference-time action
prediction depends only on observations available at the current
control step. The RGB, tactile, and action branches are jointly
optimized with flow matching using loss weights $(1.0, 0.2, 1.0)$,
respectively. All remaining FastWAM training and inference settings
follow the official configuration and are reported in
Table~\ref{tab:baseline_training_details}.

\begin{table}[!h]
\centering
\caption{Training and inference configurations for the three baseline policy families.}
\label{tab:baseline_training_details}
\footnotesize
\setlength{\tabcolsep}{5pt}
\renewcommand{\arraystretch}{1.08}
\resizebox{\textwidth}{!}{%
\begin{tabular}{@{}llll@{}}
\toprule
\textbf{Setting} & \textbf{Diffusion Policy} &
$\boldsymbol{\pi}_{0.5}$ & \textbf{FastWAM} \\
\midrule
\multicolumn{4}{@{}l}{\textit{Model and compute}} \\
Policy backbone
& Image-conditioned diffusion U-Net & --- & --- \\
GPUs
& $1 \times$ NVIDIA A100-80GB
& $8 \times$ NVIDIA A100-80GB
& $8 \times$ NVIDIA A100-80GB \\
Precision
& --- & bfloat16 mixed precision & bfloat16 \\
Distributed training
& --- & FSDP & ZeRO-1 \\
Fine-tuning method
& --- & LoRA & --- \\
LoRA rank / alpha (backbone)
& --- & $16 / 16$ & --- \\
LoRA rank / alpha (action expert)
& --- & $32 / 32$ & --- \\
\midrule
\multicolumn{4}{@{}l}{\textit{Training and optimization}} \\
Training duration
& 20 epochs & 7k steps & 10 epochs \\
Batch size (micro / global)
& --- & $32 / 256$ & --- \\
Gradient accumulation
& 1 & 1 & 1 \\
Optimizer
& AdamW & AdamW & AdamW \\
Learning rate
& $1 \times 10^{-4}$
& $2.5 \times 10^{-5}$ (peak)
& $1 \times 10^{-4}$ \\
AdamW betas
& $(0.95, 0.999)$ & $(0.9, 0.95)$ & $(0.9, 0.95)$ \\
Weight decay
& $1 \times 10^{-6}$ & $1 \times 10^{-10}$ & $1 \times 10^{-2}$ \\
LR schedule
& Cosine decay; 100-step warmup
& Cosine decay; 1k-step warmup
& Cosine decay; 5\% warmup \\
Gradient clipping
& --- & $1.0$ & $1.0$ \\
EMA
& Enabled & Disabled & --- \\
\midrule
\multicolumn{4}{@{}l}{\textit{Closed-loop inference}} \\
Denoising timesteps (train / inference)
& DDPM: $100 / 100$ & --- & Flow matching: $1000 / 10$ \\
Action horizon
& 16 & 50 & 32 \\
Actions executed per replanning
& 8 & 10 & 10 \\
\bottomrule
\end{tabular}
}
\end{table}

Table~\ref{tab:vo_vt_variant_details} summarizes the differences
between the vision-only (VO) and visuo-tactile (VT) variants of
Diffusion Policy, $\pi_{0.5}$, and FastWAM in terms of trainable
parameters, batch sizes, and input modalities.

\begin{table}[!h]
\centering
\caption{Differences between the vision-only and visuo-tactile policy
variants. Trainable parameter counts include all optimized modules and
exclude frozen model weights and preprocessing components.}
\label{tab:vo_vt_variant_details}
\footnotesize
\setlength{\tabcolsep}{5pt}
\renewcommand{\arraystretch}{1.15}
\begin{tabular}{@{}llcc@{}}
\toprule
\textbf{Policy} & \textbf{Setting} & \textbf{Vision-only (VO)} &
\textbf{Visuo-tactile (VT)} \\
\midrule

Diffusion Policy
& Trainable parameters
& $329.2$M
& $455.7$M \\

& Batch size (micro / global)
& $256 / 256$
& $128 / 128$ \\

& Input modalities
& \shortstack{RGB; robot state}
& \shortstack{RGB; robot state;\\tactile RGB; marker motion} \\

\midrule

$\pi_{0.5}$
& Trainable parameters
& $467.0$M
& $538.0$M \\

& Batch size (micro / global)
& $32 / 256$
& $32 / 256$ \\

& Input modalities
& \shortstack{RGB; robot state; language}
& \shortstack{RGB; robot state; language;\\tactile RGB; marker motion} \\

\midrule

FastWAM
& Trainable parameters
& $6.02$B
& $7.05$B \\

& Batch size (micro / global)
& $24 / 192$
& $16 / 128$ \\

& Input modalities
& \shortstack{RGB; robot state; language}
& \shortstack{RGB; robot state; language;\\tactile RGB; marker motion} \\

\bottomrule
\end{tabular}
\end{table}

\section{Per-Condition Out-of-Distribution Results}
\label{sec:ood_per_condition}

Tables~\ref{tab:ood_dp}--\ref{tab:ood_fastwam} break the aggregate
out-of-distribution results of the main paper down into the individual
held-out conditions of Table~\ref{tab:variation_factors}, for each
policy family, input modality, and deformable suite. Following the
main paper, TSR is the fraction of episodes that complete the task,
and DSR is the fraction that complete the task \emph{and} keep
$R_{\max}\le 1$ throughout; DSR is nested inside TSR, so it can never
exceed it.

\begin{table}[!h]
\centering
\renewcommand{\arraystretch}{1.1}
\setlength{\tabcolsep}{3pt}
\small
\begin{tabular}{@{}ll rr rr rr rr@{}}
\toprule
& & \multicolumn{4}{c}{\textsc{Object-Soft}}
  & \multicolumn{4}{c}{\textsc{Spatial-Soft}} \\
\cmidrule(lr){3-6}\cmidrule(l){7-10}
& & \multicolumn{2}{c}{VO-C} & \multicolumn{2}{c}{VT-C}
  & \multicolumn{2}{c}{VO-C} & \multicolumn{2}{c}{VT-C} \\
\cmidrule(lr){3-4}\cmidrule(lr){5-6}\cmidrule(lr){7-8}\cmidrule(l){9-10}
\textbf{Factor} & \textbf{Level}
& TSR & DSR & TSR & DSR & TSR & DSR & TSR & DSR \\
\midrule
\multirow{3}{*}{Lighting} & $\times$0.5  & 0 & 0 & 1 & 1 & 0 & 0 & 19 & 14 \\
                          & $\times$1.33 & 37 & 33 & 36 & 30 & 18 & 15 & 28 & 16 \\
                          & $\times$2.0  & 0 & 0 & 0 & 0 & 16 & 9 & 17 & 13 \\
\midrule
\multirow{3}{*}{Mass}     & $\times$1.25 & 41 & 38 & 42 & 34 & 5 & 5 & 18 & 17 \\
                          & $\times$1.75 & 40 & 35 & 43 & 38 & 6 & 6 & 18 & 17 \\
                          & $\times$2.5  & 36 & 33 & 33 & 27 & 5 & 5 & 17 & 15 \\
\midrule
\multirow{3}{*}{Young's}  & $\times$0.5  & 33 & 32 & 41 & 30 & 19 & 13 & 38 & 23 \\
                          & $\times$0.8  & 37 & 31 & 45 & 33 & 17 & 16 & 34 & 20 \\
                          & $\times$2.0  & 39 & 37 & 39 & 32 & 13 & 10 & 37 & 25 \\
\bottomrule
\end{tabular}
\caption{Per-condition OOD results for Diffusion Policy (\%). Each
policy--suite configuration contains 900 episodes in total: 100 per
condition and 10 per task.}
\label{tab:ood_dp}
\end{table}

\begin{table}[!h]
\centering
\renewcommand{\arraystretch}{1.1}
\setlength{\tabcolsep}{3pt}
\small
\begin{tabular}{@{}ll rr rr rr rr@{}}
\toprule
& & \multicolumn{4}{c}{\textsc{Object-Soft}}
  & \multicolumn{4}{c}{\textsc{Spatial-Soft}} \\
\cmidrule(lr){3-6}\cmidrule(l){7-10}
& & \multicolumn{2}{c}{VO-C} & \multicolumn{2}{c}{VT-C}
  & \multicolumn{2}{c}{VO-C} & \multicolumn{2}{c}{VT-C} \\
\cmidrule(lr){3-4}\cmidrule(lr){5-6}\cmidrule(lr){7-8}\cmidrule(l){9-10}
\textbf{Factor} & \textbf{Level}
& TSR & DSR & TSR & DSR & TSR & DSR & TSR & DSR \\
\midrule
\multirow{3}{*}{Lighting} & $\times$0.5  & 37 & 34 & 39 & 35 & 30 & 24 & 38 & 33 \\
                          & $\times$1.33 & 37 & 34 & 40 & 33 & 29 & 25 & 33 & 28 \\
                          & $\times$2.0  & 38 & 35 & 43 & 34 & 28 & 22 & 31 & 26 \\
\midrule
\multirow{3}{*}{Mass}     & $\times$1.25 & 37 & 37 & 44 & 40 & 13 & 13 & 18 & 16 \\
                          & $\times$1.75 & 31 & 31 & 42 & 35 & 18 & 14 & 17 & 13 \\
                          & $\times$2.5  & 26 & 26 & 36 & 29 & 14 & 11 & 17 & 14 \\
\midrule
\multirow{3}{*}{Young's}  & $\times$0.5  & 34 & 31 & 41 & 31 & 29 & 18 & 36 & 26 \\
                          & $\times$0.8  & 45 & 40 & 42 & 33 & 29 & 22 & 31 & 26 \\
                          & $\times$2.0  & 38 & 32 & 42 & 37 & 29 & 26 & 35 & 28 \\
\bottomrule
\end{tabular}
\caption{Per-condition OOD results for $\pi_{0.5}$ (\%). Each
policy--suite configuration contains 900 episodes in total: 100 per
condition and 10 per task.}
\label{tab:ood_pi05_supp}
\end{table}

\begin{table}[!h]
\centering
\renewcommand{\arraystretch}{1.1}
\setlength{\tabcolsep}{3pt}
\small
\begin{tabular}{@{}ll rr rr rr rr@{}}
\toprule
& & \multicolumn{4}{c}{\textsc{Object-Soft}}
  & \multicolumn{4}{c}{\textsc{Spatial-Soft}} \\
\cmidrule(lr){3-6}\cmidrule(l){7-10}
& & \multicolumn{2}{c}{VO-C} & \multicolumn{2}{c}{VT-C}
  & \multicolumn{2}{c}{VO-C} & \multicolumn{2}{c}{VT-C} \\
\cmidrule(lr){3-4}\cmidrule(lr){5-6}\cmidrule(lr){7-8}\cmidrule(l){9-10}
\textbf{Factor} & \textbf{Level}
& TSR & DSR & TSR & DSR & TSR & DSR & TSR & DSR \\
\midrule
\multirow{3}{*}{Lighting} & $\times$0.5  & 63 & 63 & 54 & 54 & 30 & 29 & 39 & 37 \\
                          & $\times$1.33 & 62 & 60 & 57 & 57 & 33 & 33 & 48 & 47 \\
                          & $\times$2.0  & 47 & 45 & 56 & 56 & 33 & 33 & 45 & 45 \\
\midrule
\multirow{3}{*}{Mass}     & $\times$1.25 & 59 & 59 & 61 & 61 & 17 & 17 & 21 & 21 \\
                          & $\times$1.75 & 46 & 46 & 55 & 55 & 16 & 16 & 24 & 24 \\
                          & $\times$2.5  & 38 & 38 & 49 & 49 & 16 & 16 & 23 & 23 \\
\midrule
\multirow{3}{*}{Young's}  & $\times$0.5  & 58 & 58 & 56 & 56 & 36 & 35 & 53 & 52 \\
                          & $\times$0.8  & 58 & 58 & 58 & 58 & 29 & 28 & 53 & 53 \\
                          & $\times$2.0  & 58 & 58 & 57 & 57 & 39 & 38 & 48 & 47 \\
\bottomrule
\end{tabular}
\caption{Per-condition OOD results for FastWAM (\%). Each
policy--suite configuration contains 900 episodes in total: 100 per
condition and 10 per task.}
\label{tab:ood_fastwam}
\end{table}

\section{Additional Analyses}
\label{sec:analyses}

\paragraph{Qualitative comparison of scored rollouts.}
\begin{figure}[p]
    \centering
    \includegraphics[width=0.94\linewidth,height=0.80\textheight,
    keepaspectratio]{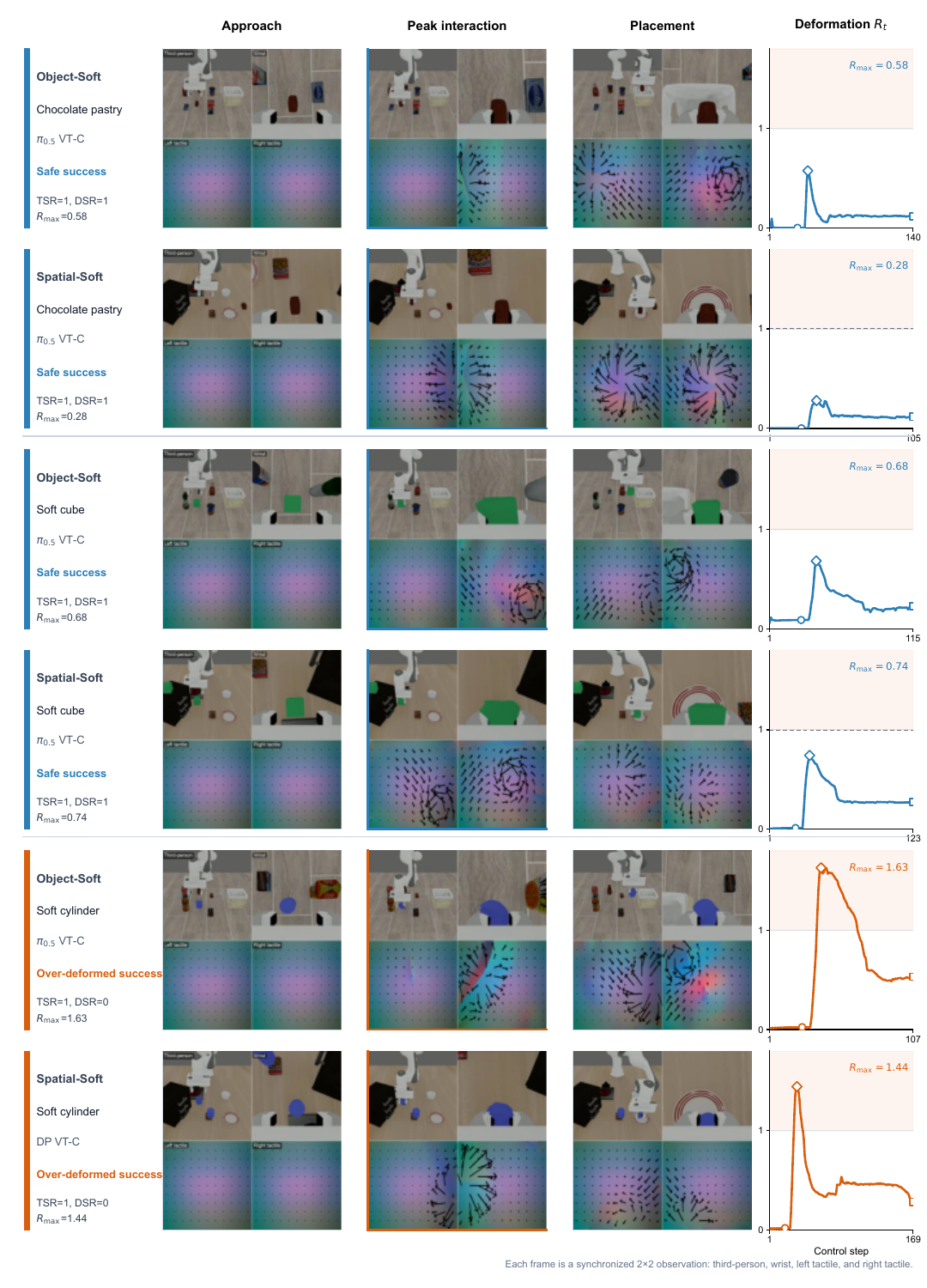}
\caption{
Illustrative rollouts on three deformable objects across
\textsc{Object-Soft} and \textsc{Spatial-Soft}. Each row shows synchronized
third-person, wrist, and tactile observations at approach, peak interaction,
and placement, followed by the normalized deformation trace $R_t$.
Circle, diamond, and square indicate the displayed steps. All rollouts
achieve task success (TSR$=1$). Blue trajectories satisfy the calibrated
deformation tolerance (DSR$=1$), while orange trajectories exceed $R_t=1$
despite success (DSR$=0$). Rows are independently sampled and not
matched-initial-state comparisons.
}
    \label{fig:qualitative}
\end{figure}
Figure~\ref{fig:qualitative} illustrates the distinction exposed by
process-level evaluation: all displayed policies ultimately place the
object successfully, but two trajectories cross the calibrated
object-specific deformation limit during interaction. The curve and the
three visual observations in each row come from the same synchronized
rollout, and the peak-interaction frame is selected at $\arg\max_t R_t$.
The tactile marker fields make the contact evolution visible, while the
DSR label itself is computed from the FEM-based deformation trace. Because
the rows cover different tasks and policy configurations, they provide
qualitative coverage rather than a controlled comparison of policy
performance. The cases were selected to span three object geometries and
both DSR outcomes; they are not used to estimate outcome prevalence.

\paragraph{Deformation measurement under stiffness shift.}
Table~\ref{tab:stiffness} reports TSR and DSR alongside the median and
interquartile range of $R_{\max}$ over the complete 100-episode paired
subset for each policy, input modality, and stiffness level. The two
axes move independently, and in both directions. At nominal stiffness,
the two Diffusion Policy variants complete equally often (TSR 42 for
both) while differing by 9 points in DSR and by 0.06 in median
$R_{\max}$, so deformation separates configurations that completion
ranks as equal. Conversely, softening the object to $\times0.5$ lowers
the completion rate of vision-only Diffusion Policy from 42 to 33 while
its median $R_{\max}$ barely moves (0.63 to 0.66), so completion also
registers effects that the deformation median does not.
Because these are closed-loop rollouts in which the policy may adapt
its actions to the perturbed dynamics, we claim only that the two axes
are non-redundant under stiffness shift; we do not claim that either
axis tracks the other, nor that $R_{\max}$ varies monotonically with
stiffness.

\begin{table}[!h]
\centering
\renewcommand{\arraystretch}{1.1}
\setlength{\tabcolsep}{4pt}
\scriptsize
\begin{tabular}{@{}lllcccc@{}}
\toprule
\textbf{Policy} & \textbf{$E$ scale} & \textbf{Input} & TSR & DSR &
\shortstack{$R_{\max}$ median\\{[IQR]}} & $N$ \\
\midrule
\multirow{8}{*}{DP}
& $\times1.0$ & VO-C & 42 & 39 & 0.63 [0.47, 0.75] & 100 \\
&              & VT-C & 42 & 30 & 0.69 [0.48, 0.97] & 100 \\
& $\times0.5$ & VO-C & 33 & 32 & 0.66 [0.06, 0.76] & 100 \\
&              & VT-C & 41 & 30 & 0.76 [0.57, 0.98] & 100 \\
& $\times0.8$ & VO-C & 37 & 31 & 0.63 [0.06, 0.79] & 100 \\
&              & VT-C & 45 & 33 & 0.75 [0.54, 0.96] & 100 \\
& $\times2.0$ & VO-C & 39 & 37 & 0.60 [0.37, 0.71] & 100 \\
&              & VT-C & 39 & 32 & 0.63 [0.41, 0.92] & 100 \\
\midrule
\multirow{8}{*}{$\pi_{0.5}$}
& $\times1.0$ & VO-C & 45 & 42 & 0.46 [0.15, 0.60] & 100 \\
&              & VT-C & 41 & 35 & 0.63 [0.51, 0.88] & 100 \\
& $\times0.5$ & VO-C & 34 & 31 & 0.56 [0.35, 0.83] & 100 \\
&              & VT-C & 41 & 31 & 0.71 [0.55, 0.92] & 100 \\
& $\times0.8$ & VO-C & 45 & 40 & 0.50 [0.31, 0.69] & 100 \\
&              & VT-C & 42 & 33 & 0.68 [0.54, 0.96] & 100 \\
& $\times2.0$ & VO-C & 38 & 32 & 0.46 [0.27, 0.64] & 100 \\
&              & VT-C & 42 & 37 & 0.59 [0.47, 0.79] & 100 \\
\midrule
\multirow{8}{*}{FastWAM}
& $\times1.0$ & VO-C & 64 & 64 & 0.49 [0.38, 0.59] & 100 \\
&              & VT-C & 58 & 58 & 0.50 [0.39, 0.57] & 100 \\
& $\times0.5$ & VO-C & 58 & 58 & 0.55 [0.42, 0.63] & 100 \\
&              & VT-C & 56 & 56 & 0.54 [0.44, 0.60] & 100 \\
& $\times0.8$ & VO-C & 58 & 58 & 0.49 [0.40, 0.62] & 100 \\
&              & VT-C & 58 & 58 & 0.50 [0.41, 0.57] & 100 \\
& $\times2.0$ & VO-C & 58 & 58 & 0.47 [0.33, 0.58] & 100 \\
&              & VT-C & 57 & 57 & 0.48 [0.39, 0.57] & 100 \\
\bottomrule
\end{tabular}
\caption{Completion and deformation statistics under stiffness shift
on \textsc{Object-Soft} (rates in \%).}
\label{tab:stiffness}
\end{table}

\section{Tactile Simulation Pipeline and Scope}
\label{sec:tactile}

\paragraph{Pipeline.}
Tactile observations are produced by TacEx~\cite{nguyen2024tacex},
which integrates GelSight-style sensor simulation into Isaac Sim. At
each control step, the gel--object interaction is derived from the
simulator contact state; Taxim~\cite{si2022taxim} renders the optical
tactile image from the resulting contact geometry using its
example-based calibration approach, and FOTS~\cite{zhao2024fots}
produces the marker-motion field. Each finger yields a
$320{\times}240$ tactile RGB image and an $11{\times}9$ marker field;
both fingers are recorded at the 20\,Hz control rate, synchronized
with all other streams.
SoftVTBench retains the TacEx GelSight Mini rendering profile and its
Taxim example-based optical calibration rather than fitting a new
real-sensor calibration. Markers are arranged as a uniform
$11{\times}9$ field (99 markers per finger); the released observations
store both the rendered RGB contact image and the 2D displacement of
every marker. No benchmark-specific physical GelSight calibration is
used.

\paragraph{Scope.}
The component simulators were validated in their original publications
under their respective conditions: Taxim reports real-to-sim accuracy
against a physical GelSight sensor on its calibration objects, and
FOTS evaluates marker-motion realism for sim-to-real skill transfer.
These published validations concern the component methods under their
own experimental conditions; they do not constitute a sim-to-real
validation of SoftVTBench's specific assets, materials, contact
conditions, or rendering parameters. We do not claim validated
transfer of SoftVTBench's simulated tactile signals to physical
GelSight sensors, and no real-sensor comparison is performed in this
work. The benchmark's conclusions are accordingly stated for simulated
visuo-tactile policy learning; characterizing the simulation-to-real
tactile gap on these assets is left to future work.

\end{document}